\documentclass[journal,twoside,web]{ieeecolor}
\usepackage{tmi}
\usepackage{cite}
\usepackage{amsmath,amssymb,amsfonts}
\usepackage{algorithmic}
\usepackage{graphicx}
\usepackage{textcomp}
\usepackage{hyperref}[colorlinks=true, linkcolor=blue, citecolor=blue]
\usepackage{multirow}
\usepackage{booktabs}
\usepackage{environ}
\usepackage{xcolor}

\newlength{\myl}
\let\origequation=\equation
\let\origendequation=\endequation
\RenewEnviron{equation}{
  \settowidth{\myl}{$\BODY$}                       
  \origequation
  \ifdimcomp{\the\linewidth}{>}{\the\myl}
  {\ensuremath{\BODY}}                             
  {\resizebox{\linewidth}{!}{\ensuremath{\BODY}}}  
  \origendequation
}

\def\BibTeX{{\rm B\kern-.05em{\sc i\kern-.025em b}\kern-.08em
    T\kern-.1667em\lower.7ex\hbox{E}\kern-.125emX}}
\begin{document}
\title{Adaptive Conditional Contrast-Agnostic Deformable Image Registration with Uncertainty Estimation}
\author{Yinsong Wang, Xinzhe Luo, Siyi Du and Chen Qin
\thanks{Manuscript submitted for review on 2 April 2025. This work was partially supported by the Engineering and Physical Sciences Research Council under Grant EP/Y002016/1 and EP/X039277/1.}
\thanks{The authors are with the Department of Electrical and Electronic Engineering, Imperial College London,
London SW7 2AZ, UK (e-mail: y.wang23@imperial.ac.uk; x.luo@imperial.ac.uk;s.du23@imperial.ac.uk; c.qin15@imperial.ac.uk).}}

\maketitle

\begin{abstract}
Deformable multi-contrast image registration is a challenging yet crucial task due to the complex, non-linear intensity relationships across different imaging contrasts. Conventional registration methods typically rely on iterative optimization of the deformation field, which is time-consuming.
Although recent learning‐based approaches enable fast and accurate registration during inference, their generalizability remains limited to the specific contrasts observed during training. In this work, we propose an adaptive conditional contrast‐agnostic deformable image registration framework (AC-CAR) based on a random convolution-based contrast augmentation scheme. AC-CAR can generalize to arbitrary imaging contrasts without observing them during training. To encourage contrast-invariant feature learning, we propose an adaptive conditional feature modulator (ACFM) that adaptively modulates the features and the contrast-invariant latent regularization to enforce the consistency of the learned feature across different imaging contrasts. Additionally, we enable our framework to provide contrast-agnostic registration uncertainty by integrating a variance network that leverages the contrast-agnostic registration encoder to improve the trustworthiness and reliability of AC-CAR. Experimental results demonstrate that AC-CAR outperforms baseline methods in registration accuracy and exhibits superior generalization to unseen imaging contrasts. Code is available at https://github.com/Yinsong0510/AC-CAR.

\end{abstract}


\section{Introduction}
\label{sec:introduction}
\IEEEPARstart{M}{ulti-contrast} deformable image registration establishes dense spatial correspondences that align images with different imaging contrasts, which is crucial for downstream multi-contrast analysis and interpretation. Precise anatomical alignment between multi-contrast images provides complementary information for characterizing tissues of human bodies, which are widely used in both qualitative and quantitative imaging in clinical diagnosis \cite{alam2018medical}. 
However, due to the complex and non-linear relationship between their intensity distributions, registering images across imaging contrasts is extremely challenging.

Although there are many conventional approaches \cite{sotiras2013deformable} on deformable image registration based on iterative optimization, most of them focus on mono-contrast scenarios and adopt intensity-based similarity measures like Mean Squared Error (MSE), which are not feasible in multi-contrast registration scenarios. To tackle this, previous multi-contrast registration approaches have leveraged information-theoretic similarity measures like mutual information \cite{viola1997alignment,maes1997multimodality}, or structural feature descriptors like MIND \cite{heinrich2012mind,heinrich2013towards} on misaligned image pairs to guide the registration. 

However, the iterative-based optimization framework is usually time-consuming and therefore could be limited in real-time applications.
Recently, deep learning-based registration approaches have shown their great potential for fast and accurate registration in both mono-contrast scenraio \cite{cheng2024winet,qin2018joint} and multi-contrast scenraio \cite{qin2019unsupervised,dey2022contrareg}. For learning-based multi-contrast registration, MIDIR \cite{qiu2021learning} and ContraReg \cite{dey2022contrareg} managed to embed information-theoretic similarity measures or their lower bound into deep neural networks. Another type of approach, like UMDIR \cite{qin2019unsupervised} and Arar et al. \cite{Arar_2020_CVPR} proposed to reduce the multi-contrast registration problem into a mono-contrast one based on image-to-image translation. 
Nevertheless, these approaches can only deal with fixed contrasts that have seen during training, resulting in poor generalizability.

Some recent work attempted to address the contrast variations between training and inference. One kind of such work \cite{hoffmann2021synthmorph} tried to achieve contrast-agnostic registration by synthesizing images of arbitrary contrasts based on randomly generated labels or segmentation maps. However, it relies on the availability of segmentation maps or the quality of generated labels. Alternatively, several recent works proposed to train a neural network to learn a certain distance metric for multi-contrast registration \cite{sideri2023mad,ronchetti2023disa}.
Nevertheless, this kind of approach focused on similarity approximation without explicitly enforcing contrast-agnostic learning.

In this work, we propose an adaptive conditional contrast-agnostic deformable image registration framework (AC-CAR) that requires only single contrast images for training but can be generalized to unseen contrasts during inference. Specifically, we propose a random convolution-based contrast augmentation scheme to generate arbitrary contrast images as input during training from single-contrast images. To learn contrast-invariant feature representations from the arbitrary contrast images generated, we propose an adaptive conditional feature modulator (ACFM). ACFM utilizes low-frequency image information to adaptively modulate feature representation to inversely remove contrast-related information. To further enhance such modulation to capture contrast-invariant features, we propose contrast-invariant latent regularization via a contrast-invariance loss. This regularizes the latent space to enforce consistent feature representations of the images with similar structures but varying imaging contrasts.
In addition, AC-CAR can rely on simple mono-contrast similarity losses, which enable network training to be more efficient and reliable. 
To equip AC-CAR with the capability to estimate contrast-agnostic uncertainty, we further integrate a separate contrast-agnostic variance network to estimate the variance map as the registration uncertainty for multi-contrast images. To the best of our knowledge, this has not been explored before, whereas it can be crucial for informing reliability and for further downstream analysis. We validate AC-CAR on T1 and T2-weighted brain MRI and cardiac T1 mapping data with different T1 weightings, which show that AC-CAR outperforms existing SOTA methods across various contrasts, even when the model has not seen most of those contrasts during training. Specifically, our contribution can be summarized as follows:

\begin{itemize}
\item[1)] We propose a new adaptive conditional contrast-agnostic registration framework based on contrast augmentation that can generalize beyond the contrasts seen during training.
\item[2)] We propose a novel adaptive conditional feature modulator to adaptively modulate the features of varying contrasts coupled with a contrast invariance loss to learn contrast-invariant features.
\item[3)] We equip the proposed registration framework with a contrast-agnostic uncertainty estimation mechanism to provide registration uncertainty.
\item[4)] Experiments show that AC-CAR outperforms the state-of-the-art baseline methods in terms of registration accuracy while providing meaningful uncertainty estimation across various contrasts.
\end{itemize}

\section{Related Works}\label{sec:review}
\subsection{Conventional Multi-contrast Image Registration}
Conventional multi-contrast deformable image registration approaches optimize the deformation field iteratively based on certain energy functions for each image pair. The problem can be formulated as follows:
\begin{equation}
	\phi^{\ast} = \underset{\phi}{\operatorname{argmin}}\left\{\mathcal{L}_{\text{sim}}(I_m \circ \phi, I_f) + \lambda\mathcal{L}_{\text{reg}}(\phi)\right\},
\end{equation}
where $I_m$ and $I_f$ denote the moving and fixed images defined over an n-D spatial domain $\Omega\in\mathbb{R}^n$. $\phi^{\ast}$ is the desired optimal deformation field. The objective function consists of a similarity term $\mathcal{L}_{\text{sim}}$ and a regularization term $\mathcal{L}_{\text{reg}}$, where $\lambda$ controls the trade-off between the two terms. However, intensity-based similarity measures like Mean Squared Error (MSE) often fail in multi-contrast registration due to the intensity matching criteria being violated by the distribution shifts across imaging contrast. Local Normalized Cross Correlation (LNCC) performs better due to it relaxes the matching criteria to linear correlation. To better deal with multi-contrast images, Maes et al. \cite{maes1997multimodality,viola1997alignment} proposed to use mutual information (MI) as the similarity to measure the statistical co-occurrence of image intensity values that is robust across different contrasts. However, MI is not sensitive to local variations as it focuses on histogram matching and lacks spatial information. Previous works tried to mitigate this by incorporating high-order MI \cite{rueckert2000non} or combining spatial information \cite{pluim2000image,zhuang2011nonrigid} within MI. Alternatively, Heinrich et al. \cite{heinrich2012mind} utilized the concept of self-similarity by proposing a Modality Invariant Neighborhood Descriptor (MIND) to extract distinctive structural features in a local neighborhood that are preserved across modalities. However, such structural descriptors are still an approximation of image similarities and may not be discriminative enough for subtle anatomical structures \cite{mok2024modality}.

\subsection{Learning-based Multi-Contrast Image Registration}

Recently, deep learning has revolutionized medical image registration for its efficiency and precision. Deep learning-based image registration approaches estimate the deformation field $\phi$ through a neural network parameterized by $\theta$, i.e,  $ \phi(\theta)=\mathit{f}_{\theta}(I_m,I_f) $.
Instead of optimizing the deformation field directly, learning-based methods optimize the network parameters. After training, the network can estimate the deformation field with a single forward pass at the inference stage. Many deep learning approaches have adapted conventional multi-contrast metrics to learning-based frameworks. For example, Qiu et al. \cite{qiu2021learning} extended conventional MI using a differentiable Parzen window \cite{viola1997alignment} for network backpropagation. 


In contrast, other works addressed multi-contrast registration from perspectives beyond
conventional loss. For instance, Dey et al. \cite{dey2022contrareg} employed contrastive learning with a pre-trained autoencoder to learn an image distance between warped moving and fixed images using a multi-scale PatchNCE loss, which can be regarded as a lower bound of MI and thus inherits its limitations \cite{mok2024modality}. Alternatively, some researchers tried to use image translation-based methods to reduce the multi-contrast registration problem to a mono-contrast one. Qin et al. \cite{qin2019unsupervised} proposed to use a GAN-based image-to-image translation framework to disentangle multi-contrast images into shape and appearance. The translated image can be reconstructed by the shape code of the source image and the appearance code of the target image. The translated image thus has the same contrast as the target image, reducing the multi-contrast registration to a mono-contrast one. Similarly, Arar et al. \cite{Arar_2020_CVPR} proposed a geometry-preserving image translation network. They directly learned the translated image with L1 loss without disentanglement and iteratively exchanged the order of registration and translation during training to encourage geometry preservation. Deng et al. \cite{deng2023interpretable} disentangled multi-contrast images into features responsible for alignment (RA features) and not responsible for alignment (nRA features) by convolutional sparse encoding and used RA features solely for registration.
However, all of the approaches mentioned above can only deal with fixed imaging contrasts, which means they can not generalize to contrasts not seen at the training stage.

\subsection{Contrast-Agnostic Image Registration}
To enable the registration network to be more generalizable to unseen imaging contrasts. Hoffmann et al. \cite{hoffmann2021synthmorph} proposed synthesizing arbitrary contrast of images based on segmentation or randomly generated labels and optimized registration network by Dice loss \cite{dice1945measures}. However, the employed Dice loss mainly focuses on the structure overlaps rather than pixel-level or volume-level differences. Several recent works proposed training a neural network to learn a certain distance metric for multi-modal registration. 
Sideri et al. \cite{sideri2023mad} relied on modality and affine geometric augmentation for learning a contrast-agnostic distance measure based on image patch centers to tackle multi-contrast rigid registration.
Ronchetti et al. \cite{ronchetti2023disa} proposed to use a small CNN to learn a feature-level distance that approximates the conventional Linear Correlation of Linear Combination (LC$^2$) similarity. Mok et al. \cite{mok2024modality} proposed to learn a modality-agnostic structural representation by adapting the structural descriptor \cite{heinrich2012mind} into a CNN. The learned structural representations were then used as the input of the registration network. However, this approach requires training two sub-networks, which demands high computational costs and may impede the registration network as its performance highly depends on how well the structural representations are learned.

Of particular relevance to this work, our earlier work \cite{wang2024car} proposed a contrast-agnostic registration framework by simulating arbitrary contrast for training and regularizing the feature in latent space. In this work, to better handle the variability among different contrasts, we extend previous approaches by introducing an adaptive conditional feature modulator to adaptively modulate the features of varying contrast to capture contrast-invariant features. Additionally, a contrast-agnostic uncertainty estimation framework is introduced to equip the model with the capability to estimate uncertainty during registration. More comprehensive quantitative and qualitative evaluations of AC-CAR, including comparisons, generalization, and ablation studies, were conducted on a 3D brain MRI dataset and a 2D cardiac MRI dataset.

\begin{figure*}[htbp]
	\centering
\includegraphics[width=0.85\textwidth]{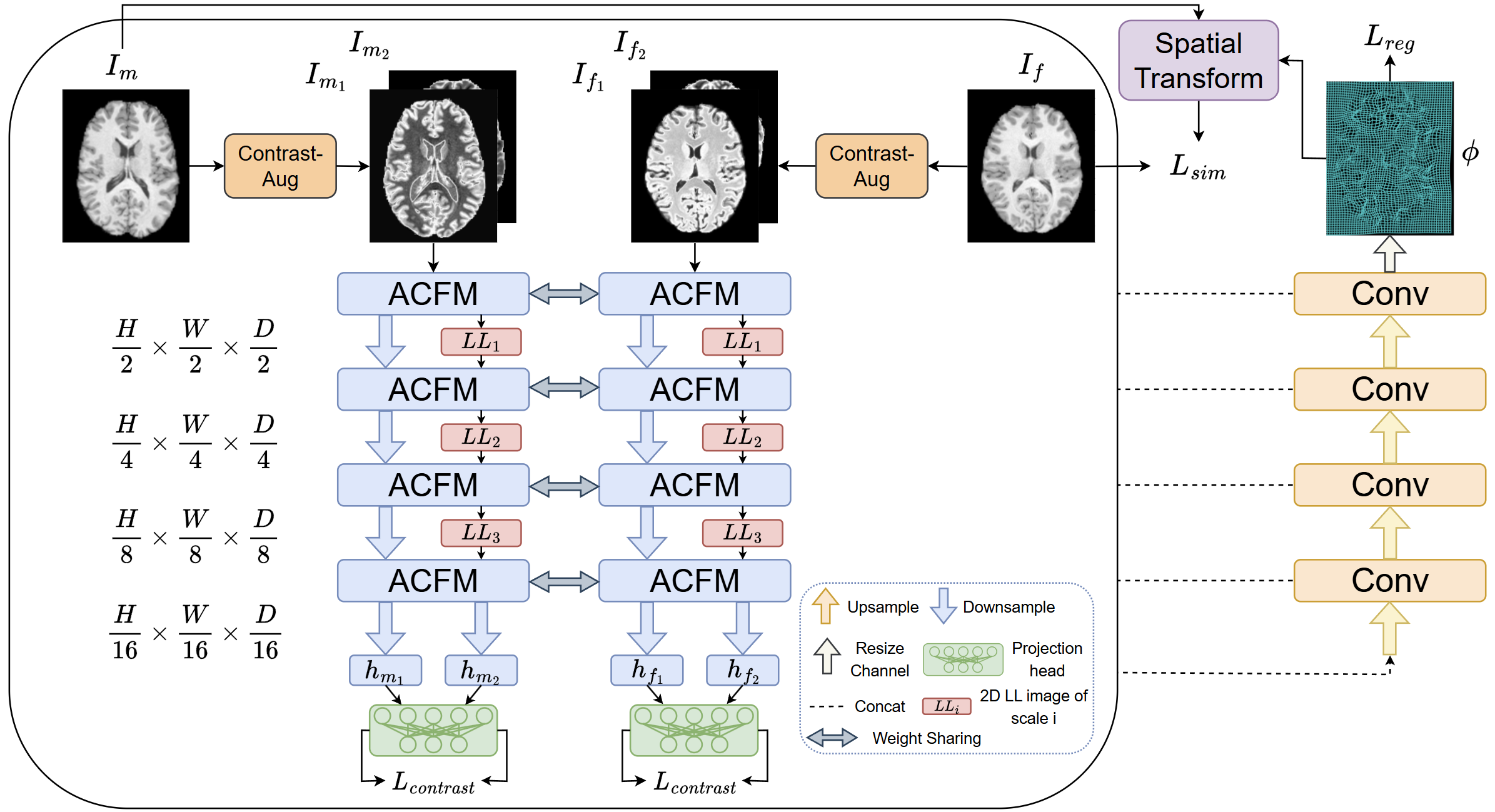} 
	\caption{Overview of the proposed registration framework. We first simulate two augmented images for each moving and fixed image with the contrast augmentation module. The augmented images are then used as the network inputs. At the encoder part, we use our proposed adaptive conditional feature modulator (ACFM) and contrast-invariant latent regularization (CLR) to extract contrast-invariant features. The learned features are then used to estimate the deformation field. The deformation field is then used to warp the pre-augmented moving image. The similarity loss is calculated on the warped image and the pre-augmented fixed image.} 
	\label{figure 3.1}
\end{figure*}

\subsection{Uncertainty Estimation in Image Registration}

Precise quantification of registration uncertainty is essential to mitigate the risk of misdiagnosis resulting from erroneous image alignments. Conventional methods \cite{risholm2013bayesian,sedghi2019probabilistic} employed probabilistic image registration (PIR) to estimate distributions of transformation parameters. The registration uncertainty can be derived from the variance of distributions. However, this approach only reflects variability in the mean model prediction, which may not correlate with registration error. 

Recently, uncertainty estimation has gained growing interest in learning-based image registration. Dalca et al. \cite{dalca2019unsupervised} integrated PIR-based uncertainty estimation into a learning-based framework. Although it benefits from fast inference of learning-based methods, it still inherits the limitations of conventional methods. To address this, Chen et al. \cite{chen2022transmorph} proposed to calibrate the registration uncertainty using the fixed image to replace the mean model prediction. Furthermore, Zhang et al. \cite{zhang2024heteroscedastic} proposed heteroscedastic uncertainty estimation by using a separate variance network to predict the variance map between the warped moving image and the fixed image. Nevertheless, existing learning-based approaches are still limited to mono-contrast uncertainty estimation, and there is still a gap in uncertainty estimation for learning-based multi-contrast image registration.

\section{Methodology}\label{sec:method}
In this work, we propose an adaptive conditional contrast-agnostic registration framework that can deal with even unobserved contrasts. Specifically, we proposed a random-convolution-based contrast-augmentation scheme that can simulate arbitrary contrast of images as inputs (Section \ref{sec3.1}). To enable the network to learn contrast-invariant features for contrast-agnostic registration, we propose an adaptive conditional feature modulator to modulate learned features (Section \ref{sec3.2}) and a contrast invariance loss to enforce consistency among features of the similar structure (Section \ref{sec3.3}). Additionally, an uncertainty estimation mechanism is incorporated into the AC-CAR for providing uncertainty estimation for contrast-agnostic registration (Section \ref{sec3.4}).

\subsection{Contrast Augmentation via Random Convolution}\label{sec3.1}
To enable a contrast-agnostic registration framework, the network needs to learn contrast-invariant features for downstream deformation prediction. However, most existing learning-based multi-contrast registration approaches \cite{qiu2021learning,dey2022contrareg} fail to generalize to unseen imaging contrasts during inference due to their dependence on a fixed set of imaging contrasts during training, leading the network to learn contrast-dependent features. This leads to a strong dependency between the input imaging contrast and its learned feature, therefore resulting in reduced generalizability.

To mitigate this, we propose a contrast augmentation scheme that simulates diverse imaging contrasts using Random Convolution (RC), similar to prior work such as \cite{sideri2023mad}. The contrast augmentation scheme stacks multiple randomly initialized convolutional kernels with kernel size equal to one and LeakyReLU activations to simulate non-linear mapping among different MRI contrasts. We follow the same network setting for the contrast augmentation scheme as our earlier proposed work \cite{wang2024car}.
The augmented images capture diverse contrast patterns, simulating potential distributions of unseen contrasts. They are then used as inputs to train our contrast-agnostic registration network. By varying the input imaging contrasts at every training iteration, this approach alleviates the network's reliance on contrast-specific information, enabling the network to predict the deformation field in a contrast-agnostic manner.

\subsection{Adaptive Conditional Feature Modulator}\label{sec3.2}

\begin{figure}[htbp]
	\centering
\includegraphics[width=0.49\textwidth]{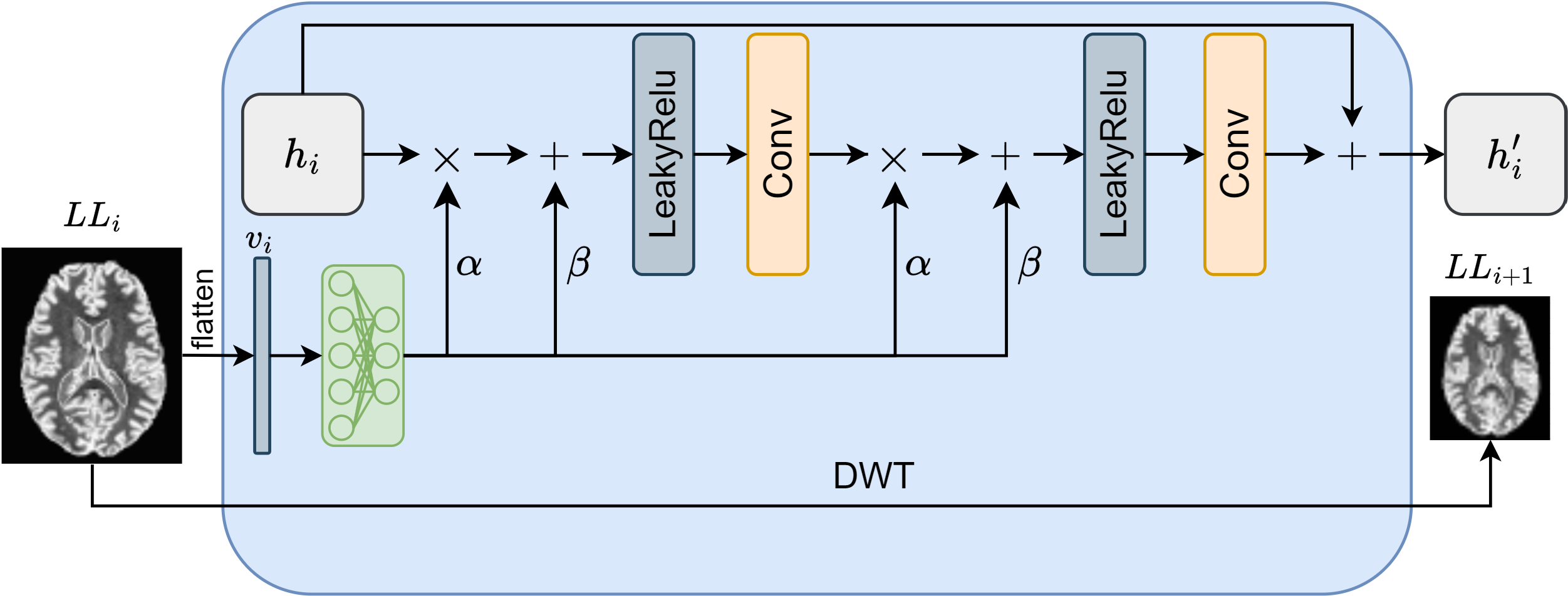} 
	\caption{Overview of the proposed Adaptive Conditional Feature Modulation Module (ACFM).} 
	\label{figure 3.2}
\end{figure}

Handling inputs of arbitrary imaging contrasts with a single encoder is challenging, as it may struggle to manage the variability in different imaging contrasts and result in a trivial solution instead of contrast-invariant feature representations. To overcome these limitations, we propose an Adaptive Conditional Feature Modulator (ACFM) that leverages conditional instance normalization (CIN) \cite{dumoulin2017a} to enable the network to adaptively modulate the feature representation according to different imaging contrasts. This is achieved by conditioning the modulation on the low-frequency component extracted via the discrete wavelet transform (DWT). Unlike previous wavelet-based registration methods such as WiNet \cite{cheng2024winet}, which use DWT to directly represent the deformation field, ACFM instead employs the DWT-extracted low-frequency component to inversely remove contrast-related variations from the learned features through CIN. Specifically, ACFM is conditioned on the low-frequency component of the input image, based on the assumption that contrast-related information is predominantly concentrated in low-frequency bands. The details of the ACFM are shown in Fig.~\ref{figure 3.2}. Within the ACFM, we first extract the low-frequency components ($LL_i$) of the image by using DWT to modulate the feature representation in a multi-scale manner, where $i$ denotes the current scale. The first ACFM is conditioned on the original image $LL_0$ at the original scale. Each following ACFM is conditioned on the low-frequency components of the conditioned image from the previous ACFM by performing another DWT. Each conditioned image of ACFM shares the same size as the feature of that layer, allowing it to modulate the feature for each scale. To use the $LL_i$ for CIN, we first expand $LL_i$ into a vector $v_i$. The $v_i$ are then fed into a projection head of a linear layer to generate the scale and shift parameters, $\alpha$ and $\beta$. The operation of CIN is summarized as
\begin{equation}
	\mathit{h_{i}'} = \alpha_{\theta,\mathit{i}}(v_i)(\frac{\mathit{h_{i}}-\mathit{\mu(h_{i})}}{\mathit{\sigma(h_{i})}}) + \beta_{\theta,\mathit{i}}(v_i),
\end{equation}
where $h_{i}$ denotes the input of the current ACFM layer in channel $i$, $ \mathit{\mu(h_{i})} $ and $ \mathit{\sigma(h_{i})} $ are the mean and standard deviation of $ \mathit{h_{i}} $, and $\alpha_{\theta,\mathit{i}}(v_i)$ and $\beta_{\theta,\mathit{i}}(v_i)$ are the scale and shift parameters for each channel in the hidden feature map $h_{i}$. By conditioning the feature representation on the low-frequency component of the input image, we enable the network to modulate the learned feature representations adaptively to remove their contrast-related information inversely.
\subsection{Contrast-Invariant Latent Regularization (CLR)}\label{sec3.3}
The proposed ACFM (Section \ref{sec3.2}) can adaptively modulate the feature according to varying input contrasts to encourage contrast-invariant feature learning. However, this modulation is not explicitly constrained to achieve this. To further enhance contrast-invariant feature learning, we propose regularizing the feature representation in latent space using our proposed contrast invariance loss. Specifically, we aim to pull the feature representations of the images with the same shape information close in latent space under the assumption that contrast-invariant feature representations should be highly correlated with the shape information of the images. We generate two sets of contrast-augmented image pairs $\{I_{m_1}, I_{f_1}\}$ and $\{I_{m_2}, I_{f_2}\}$ based on a mono-contrast image pair $\{I_m, I_f\}$ with two separate contrast augmentation. Then, the contrast-invariant loss can be formulated as follows,
\begin{equation}
\scalebox{1.0}{$
\begin{aligned}
	{\mathcal{L}_{\text{contrast}}} = \mathbb{E}_{\Omega}\bigg[&||{\operatorname{Proj}(h_{m_1}) - \operatorname{Proj}(h_{m_2}})||_2^2 
 + \\ &||{\operatorname{Proj}(h_{f_1}) - \operatorname{Proj}(h_{f_2})}||_2^2\bigg],
\end{aligned}$}
\end{equation}
where $\operatorname{Proj}$ represents the projection head, $h_{m_1}, h_{m_2}, h_{f_1}$, and $h_{f_2}$ represent the corresponding latent features after the encoder. The projection head consists of a linear layer. The contrast invariance loss aims to pull the feature representation with the same shape as close as possible in the latent space.
 
It serves two critical purposes during training: 1) It compels the encoder to learn intrinsic structural features that are robust to variations in imaging contrasts and appearances \cite{achille2018emergence}. 2) It provides guidance for the modulations of ACFM to encourage contrast-invariant feature representations to remove contrast-related information inversely.

\subsection{Contrast-Agnostic Uncertainty Estimation Network}\label{sec3.4}
\begin{figure}[htbp]
	\centering
\includegraphics[width=0.5\textwidth]{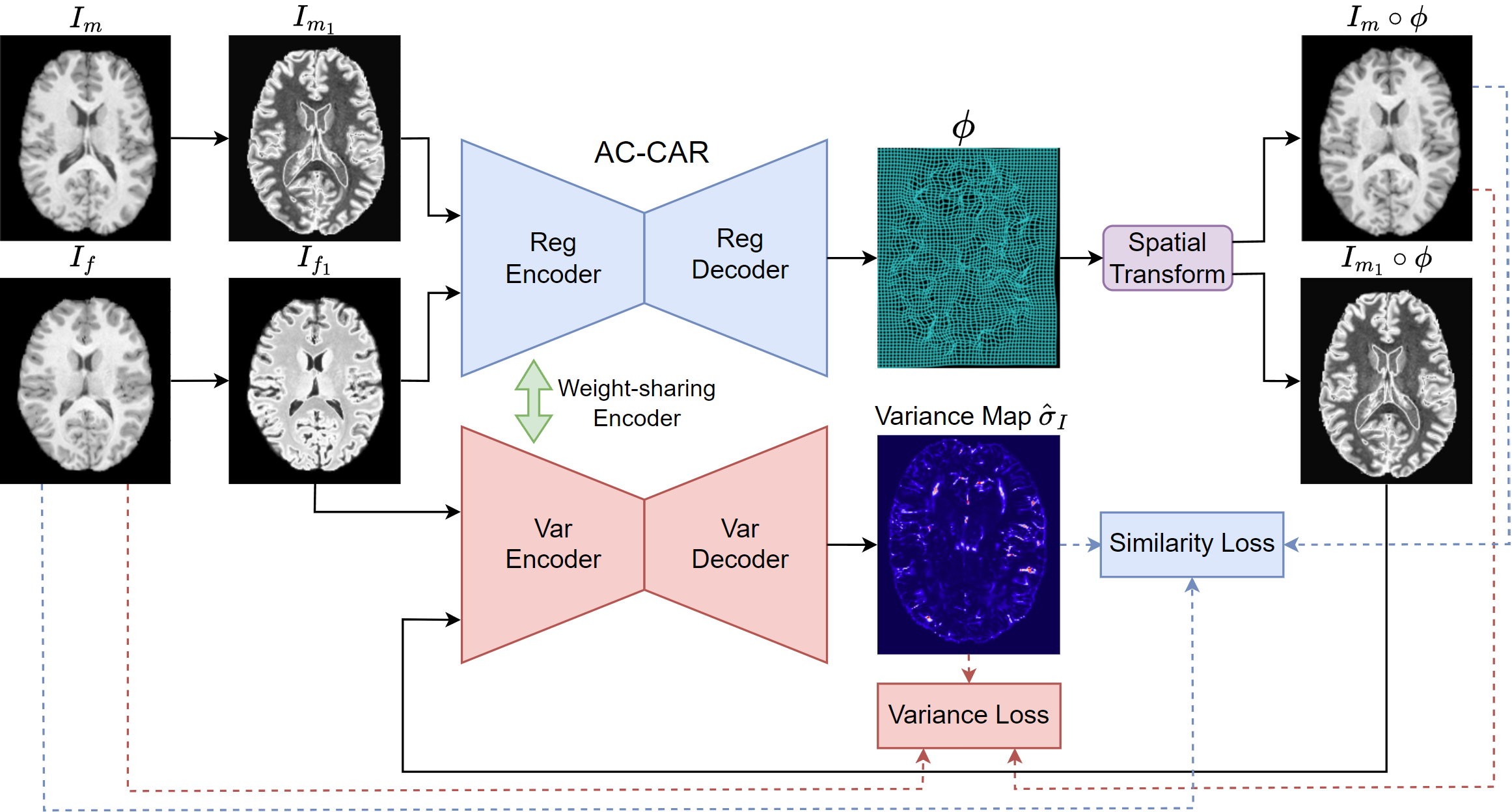} 
	\caption{Overview of the proposed contrast-agnostic uncertainty estimation framework.} 
	\label{figure 3.3}
\end{figure}

Beyond contrast-agnostic registration, we enhance the framework with a contrast-agnostic uncertainty estimation module to capture errors from heteroscedastic noise and highlight regions at higher risk of misregistration. Existing uncertainty estimation approaches for learning-based image registration are limited to mono-contrast scenarios \cite{chen2022transmorph,zhang2024heteroscedastic}. However, in multi-contrast settings, the intensity correspondence between contrasts is highly nonlinear, making it more difficult to disentangle uncertainty arising from true misalignment from that induced by contrast differences. This challenge complicates the estimation of reliable uncertainty. To address this issue, we adapt heteroscedastic uncertainty modelling to the contrast-agnostic setting using our learned contrast-invariant feature representations. Inspired by Zhang et al. \cite{zhang2024heteroscedastic}, we use a separate variance network that predicts the input-specific heteroscedastic variance map for each multi-contrast image pair. To allow the variance network to be contrast-agnostic, the variance network is constructed on the contrast-agnostic encoder of AC-CAR, which yields contrast-invariant feature representations. These features are then used to predict the variance map of the warped moving image. The network details are shown in Fig.~\ref{figure 3.3}. The variance network thus has a weight-sharing encoder with the AC-CAR network and a separate decoder, which is used to output the heteroscedastic variance map. Specifically, the variance network takes the warped moving augmented image $I_{m_1} \circ \phi$ and the fixed augmented image $I_{f_1}$ as the input. We optimize the variance network based on the $\beta$-NLL objective \cite{seitzer2022on}, similar to Zhang et al. \cite{zhang2024heteroscedastic}. Same as the registration network, we use the pre-augmented mono-contrast image pairs for loss supervision to facilitate the $\beta$-NLL objective in a multi-contrast registration scenario. The loss function for optimizing the variance network can be formulated as follows.
\begin{equation}\label{var}
\begin{aligned}
	{\mathcal{L}_{\text{var}}} = \mathbb{E}_{\Omega}\left[{\lfloor\hat{\sigma}_{I}^{2\beta}\rfloor}(\frac{1}{\hat{\sigma}_{I}^{2}}||I_m\circ\lfloor\phi \rfloor-I_f||^2_2 + \log\hat{\sigma}_{I}^{2})\right],
\end{aligned}
\end{equation}
where $\hat{\sigma}_{I}$ denotes the estimated variance map, $\beta$ denotes the hyperparameter in the $\beta$-NLL objective. $\lfloor \ \rfloor$ denotes gradient stopping to prevent duplicate back-propagation by the two subnetworks.  Similar to Zhang et al. \cite{zhang2024heteroscedastic}, we output $\log\hat{\sigma}_{I}^{2}$ for the variance network for numerical stability.

\subsection{Overall Network Architecture and Loss function}\label{sec3.5}

The overall registration framework is illustrated in Fig.~\ref{figure 3.1}. It follows a U-shaped architecture \cite{balakrishnan2019voxelmorph}, with two Siamese encoders for the moving and fixed images. Skip connections \cite{he2016deep} are employed between the two Siamese encoders and the decoder. The moving and fixed images are input into their respective encoders, producing latent features $h_{m}$ and $h_{f}$, each reduced to $1/16$ of the original image size. The two learned latent features are then concatenated and fed into the decoder, which is used to estimate the final deformation field.

The loss function of the registration network includes three terms: a similarity loss $\mathcal{L}_{\text{sim}}$, a regularization loss $\mathcal{L}_{\text{reg}}$, and the proposed contrast invariance loss $\mathcal{L}_{\text{contrast}}$. The loss function for the variance network is given in Eq.~\ref{var}. As the input contrast-augmented image pairs are generated from the mono-contrast image pair $\{I_m, I_f\}$, we propose using the pre-augmented mono-contrast image pairs for loss supervision with a mono-contrast similarity measure. This avoids designing a multi-contrast similarity loss. The loss function of the registration network can be formulated as 
\begin{equation}\label{reg}
 \begin{aligned}
	{\mathcal{L}_{\text{total}}} = &\mathbb{E}_{\Omega}\left[\lfloor\hat{\sigma}_I^{2\beta-2}\rfloor ||I_m\circ\phi-I_f||^2_2 + \lambda_1 ||\nabla\phi||_2^2\right] + \\ & \lambda_2\mathcal{L}_{\text{contrast}} - \lambda_3\operatorname{LNCC}(I_m\circ\phi,I_f),
\end{aligned} 
\end{equation}
where $\lfloor \ \rfloor$ denotes gradient stopping. $\lambda_{1}$, $\lambda_{2}$, and $\lambda_{3}$ are the respective hyperparameters of the diffusion regularizer, contrast invariance loss, and the additional LNCC loss. Similar to Zhang et al \cite{zhang2024heteroscedastic}, we adopt an alternating training strategy for the registration and variance networks.

\section{Experiments Setup}\label{sec:4}

\begin{figure*}[t!]
	\centering
\includegraphics[width=1.0\textwidth]{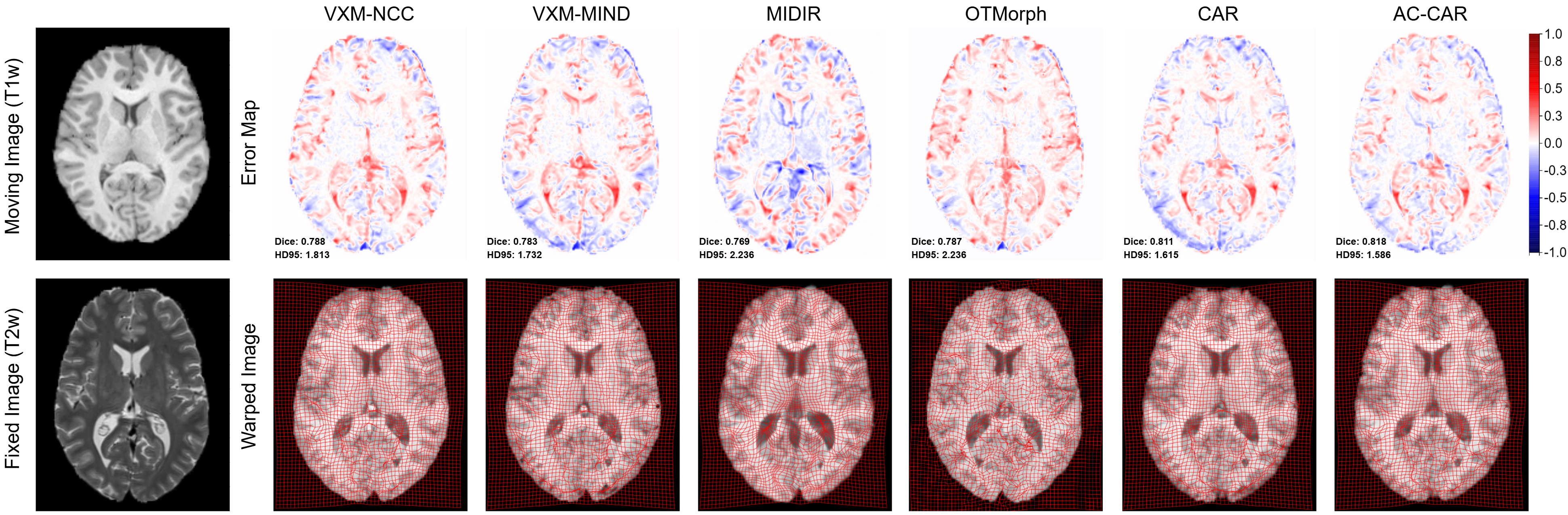} 
	\caption{Qualitative results on CamCAN dataset. We present the registration results of the middle slices along the z-axis of the same volume for illustration. The first row shows the error map of our proposed method against the baseline. The Dice score and HD95 of the whole volume are shown at the bottom left of the error map. The second row shows the warped images overlaid with the deformation fields.}
	\label{figure 4.1}
\end{figure*}

\subsection{Datasets and Preprocessing}
\subsubsection{Inter-subject 3D brain MRI registration of multi-contrast images}

For the brain registration task, we use two multi-contrast datasets. (1) The public Cambridge Centre for Ageing and Neuroscience (CamCAN) project \cite{shafto2014cambridge,taylor2017cambridge}. The dataset has 3D MRI T1-weighted (T1w) and T2-weighted (T2w) images of 652 subjects. All images have a spatial resolution of $1\times1\times1$ mm and are cropped to a size of $160\times192\times160$. The dataset was randomly split into 600, 10, and 42 subjects for training, validation, and testing, where subjects were randomly selected to form T1w-T2w image pairs. (2) The Information eXtraction from Images (IXI) database. The dataset comprises 3D MRI T1w, T2w, and Proton Density-weighted (PD) images of 397 subjects. All images are processed using Freesurfer \cite{fischl2012freesurfer} for skull-stripping. SimpleITK \cite{lowekamp2013design} is adopted for bias-field correction and affine alignment to normalize all the images into a common space and have the same spatial resolution. All images are cropped to a size of $192\times160\times144$. MALPEM \cite{ledig2015robust} is used for automated segmentation of 138 cortical and subcortical structures, categorized into 5 groups. The dataset was randomly split into 347, 10, and 40 subjects for training, validation, and testing, where subjects were randomly selected to form T1w-T2w, T1w-PD, and T2w-PD image pairs. For training AC-CAR and CAR, image pairs are formed using T1w images only for both datasets.

\subsubsection{Intra-subject registration of multiple contrasts of T1 mapping acquisition in cardiac MRI}
For the cardiac registration task, we use the public 2D CMRxRecon cardiac dataset \cite{wang2021recommendation,wang2023cmrxrecon}. CMRxRecon dataset consists of two parts: (1) Cine MRI, and (2) T1 mapping with images of nine different T1 weightings (nine different imaging contrasts, $\mathrm{TI_{i=1:9}}$). Both Cine MRI and T1 Mapping data in the dataset contain 167 subjects. Each subject includes 7 to 12 or 4 to 5 short-axis view slices for Cine MRI data and T1 Mapping data, respectively. The images in Cine and T1 Mapping data are cropped into a size of $128\times128$. The dataset was randomly split into 137, 10, and 20 subjects for training, validation, and testing. The image sequence within each slice is considered free from motion artefacts. Therefore, we simulate random deformations to distort images using artificially generated Free-Form Deformations (FFDs) with various mesh spacings following Luo et al.\cite{9965747}. To simulate realistic deformation, we restrict the deformation within the cardiac region. For sequences within the same slice, we selected the original image $\mathrm{TI_{1}}$ and a spatially randomly deformed image from $\mathrm{TI'_{i=2:9}}$ as an image pair. For training AC-CAR and CAR, the image pairs are formed by the original image $\mathrm{TI_{1}}$ and its own deformed pair $\mathrm{TI'_{1}}$.

\subsubsection{Intra-subject registration of abdominal MR-CT} 
To evaluate the performance of AC-CAR beyond multi-contrast images, we also conduct experiments on abdominal CT-MR registration using the Learn2Reg MICCAI challenge dataset \cite{hering2022learn2reg}. The Learn2Reg dataset contains 3D T1-weighted MR and CT abdominal images. The data were resampled to $3\times3\times3$ mm and cropped to a size of $112\times96\times112$. Training was conducted on 40 unpaired MR and 50 unpaired CT images, with CT-MR pairs formed through random selection. 8 CT-MR pairs were used for testing. The masks provided by the dataset were used to confine the information used for registration.  Whereas VXM-MIND and MIDIR are trained directly on CT-MR pairs, AC-CAR is trained solely on T1w MR image pairs. Informed consent was obtained during the public data release for all datasets.

\begin{figure*}[t!]
	\centering
\includegraphics[width=1.0\textwidth]{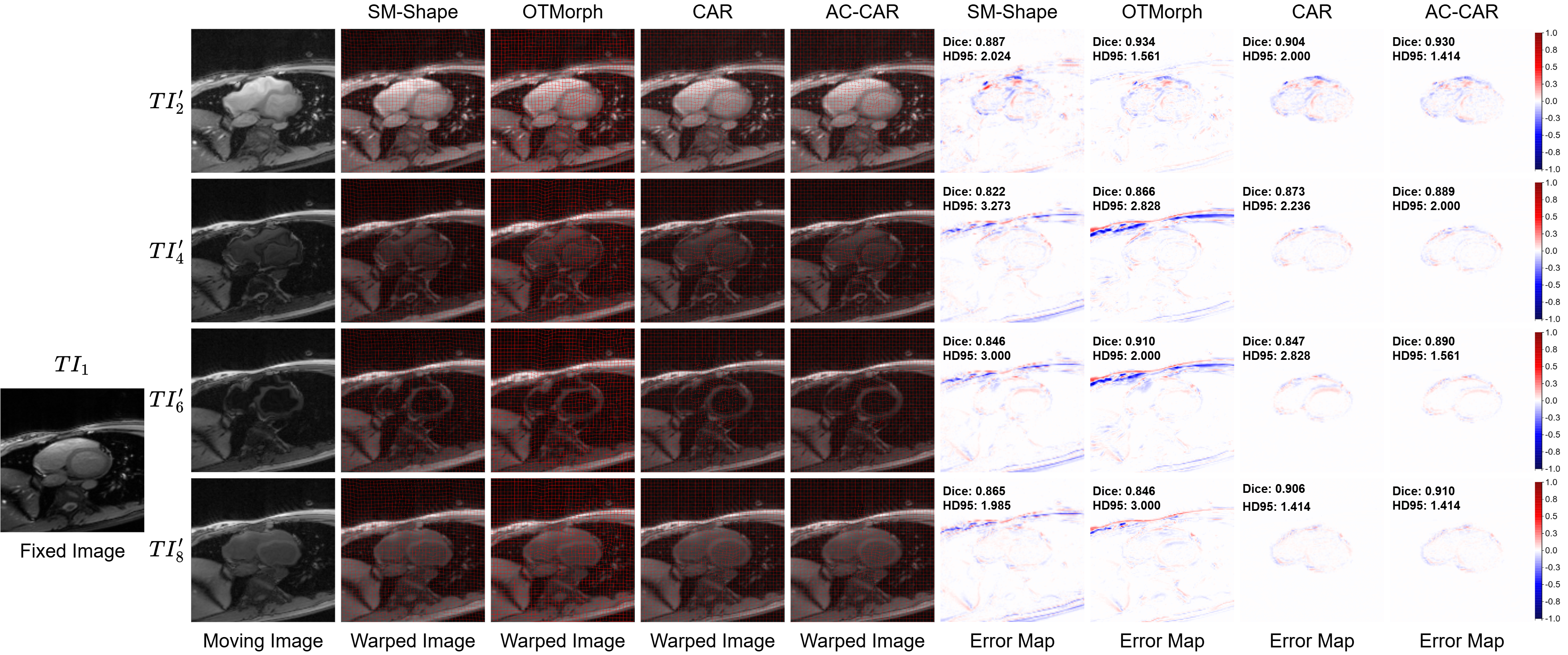} 
	\caption{Qualitative results on CMRxRecon dataset using T1 Mapping data for training. We show the results of registering misaligned images of $\mathrm{TI_{i=2,4,6,8}}$ to $\mathrm{TI_{1}}$. Columns 2-5 present the warped images overlaid with the deformation fields, and columns 6-10 present the error map of our proposed method against the baseline. The Dice score and HD95 are shown at the top left of the error map.}
	\label{figure 4.2}
\end{figure*}

\begin{table*}[!t]
\setlength{\tabcolsep}{1.5 mm}
\caption{Quantitative Results of on both CamCAN and CMRxRecon datasets. * represents for values that AC-CAR significantly outperformed with $p$-value $< 0.01$ in a paired $t$-test.}
\centering
\scalebox{0.95}{
\begin{tabular}{ccccccccccc}
  \toprule[1pt]
  \multirow{2}{*}{\textbf{Methods}} & \multicolumn{4}{c}{\textbf{CamCAN (T1w$\rightarrow$T2w)}} & \multicolumn{4}{c}{\textbf{CMRxRecon}} & \\
    \cmidrule(lr){2-9}
  
   	& Dice $\uparrow$ & $|\nabla_{J}|$ $\downarrow$ & $J_{<0}$\% $\downarrow$ & HD95 $\downarrow$ & Dice $\uparrow$ & $|\nabla_{J}|$ $\downarrow$ & $J_{<0}$\% $\downarrow$ & HD95 $\downarrow$ \\
	\cmidrule(lr){1-9}
 Unregistered & 0.609*$\pm$0.078 & - & - & 5.418*$\pm$1.829 & 0.783*$\pm$0.079 & - & - &  2.749*$\pm$0.546\\
 SyN & 0.756*$\pm$0.023 & \textbf{0.043}$\pm$0.005 & \textbf{0.430}$\pm$0.146\% & 2.033*$\pm$0.199 & 0.601*$\pm$0.271 & 0.019*$\pm$0.006 & 0.034*$\pm$0.073\% & 6.142*$\pm$4.469\\
    VXM-LNCC & 0.753*$\pm$0.029 & 0.065*$\pm$0.005 & 0.549*$\pm$0.122\% & 2.098*$\pm$0.187 & 0.838*$\pm$0.085 & 0.051*$\pm$0.015 & 0.004$\pm$0.008\% & 2.301*$\pm$0.760\\
    VXM-MIND & 0.746*$\pm$0.032 & 0.065*$\pm$0.006 & 0.619*$\pm$0.228\% & 2.157*$\pm$0.198 & 0.837*$\pm$0.077 & 0.051*$\pm$0.015 & 0.007*$\pm$0.020\% & 2.158*$\pm$0.532\\
    MIDIR & 0.761*$\pm$0.023 & 0.052$\pm$0.005 & 0.600*$\pm$0.146\% & 2.152*$\pm$0.175 & 0.795*$\pm$0.092 & 0.023*$\pm$0.005 & 0.018*$\pm$0.045\% & 2.670*$\pm$0.728\\
    SM-Shape & 0.698*$\pm$0.063 & 0.074*$\pm$0.009 & 0.513$\pm$0.112\% & 3.305*$\pm$1.764 & 0.836*$\pm$0.078 & 0.049*$\pm$0.005 & 0.006*$\pm$0.016\% & 2.302*$\pm$0.576\\
    CAR & 0.784*$\pm$0.029 & 0.063*$\pm$0.007 & 0.558*$\pm$0.145\% & 1.824*$\pm$0.239 & 0.860*$\pm$0.073 & 0.014$\pm$0.005 & 0.004$\pm$0.010\% & 2.016*$\pm$0.468\\
    OTMorph & 0.775*$\pm$0.026 & 0.072*$\pm$0.006 & 0.607*$\pm$0.211\% & 1.953*$\pm$0.188 & \textbf{0.876}$\pm$0.056 & 0.032*$\pm$0.009 & 0.004$\pm$0.006\% & 2.273*$\pm$0.681 \\
    UTSRMorph & 0.771*$\pm$0.024 & 0.062*$\pm$0.006 & 0.604*$\pm$0.145\% & 1.955*$\pm$0.188 & 0.859*$\pm$0.061 & 0.024*$\pm$0.027 & 0.005$\pm$0.010\% & 2.557*$\pm$0.781\\
    \textbf{AC-CAR (w/o NCC)} & 0.787*$\pm$0.023 & 0.059$\pm$0.007 & 0.703*$\pm$0.165\% & 1.766*$\pm$0.191 & 0.863*$\pm$0.065 & \textbf{0.012}$\pm$0.004 & \textbf{0.003}$\pm$0.009\% & 1.943*$\pm$0.456\\
    \textbf{AC-CAR (Ours)} & \textbf{0.808}$\pm$0.026 & 0.056$\pm$0.006 & 0.494$\pm$0.101\% & \textbf{1.652}$\pm$0.236 & 0.871$\pm$0.068 & 0.015$\pm$0.005 & \textbf{0.003}$\pm$0.009\% & \textbf{1.862}$\pm$0.464\\
	\bottomrule[1pt]
\end{tabular}}
\label{table:quant_cam}
\end{table*}

\subsection{Evaluation Metrics}

The registration accuracy is evaluated by the Dice score and the Hausdorff Distance. The Dice score measures the degree of overlap between the anatomical segmentation of the warped moving image and the fixed image. The Hausdorff Distance (HD) captures the maximum distance between corresponding anatomical regions in the warped moving and fixed images. To mitigate the impact of outliers, we report the 95th percentile of the HD (HD95). Additionally, we measure extreme deformations by calculating the folding ratio—the percentage of points with a negative Jacobian determinant ($J_{<0}\%$), and evaluate deformation smoothness through the magnitude of the spatial gradient of the Jacobian determinant $|\nabla_{J}|$ \cite{qiu2021learning}.

\subsection{Implementation Details}
The registration network comprises two Siamese encoders followed by a decoder. For 3D brain registration, encoder layers have 32 channels and decoder layers 64 channels; for 2D cardiac registration, encoder layers have 128 channels and decoder layers 256 channels. The projection head for CLR employs a $1\times1$ convolution with 16 output channels for 3D tasks and 32 for 2D tasks. The random convolution-based contrast augmentation uses 4 RC layers, with kernel weights sampled from a uniform distribution $U(0, 10)$, re-normalized to be zero-centered, and followed by a LeakyReLU activation with a negative slope of 0.2. Configurations of contrast augmentation are based on preliminary ablation studies from our previous work \cite{wang2024car}. Haar wavelet transform is adopted for the DWT in our proposed ACFM. To reduce computational costs while maintaining performance, a single 2D slice, the middle slice along the z-axis, is used as the condition of ACFM in 3D experiments, as it is expected to encapsulate the complete contrast information of the entire 3D image. $\lambda_1$, $\lambda_2$, and $\lambda_3$ are set to 0.15, 0.1, and 0.8 for 3D experiments and 0.3, 0.2, and 0.8 for 2D experiments, chosen through empirical tuning on a validation subset to balance deformation smoothness, registration accuracy, and CLR. The parameter $\beta$ in Eq.~\ref{var} and Eq.~\ref{reg} is fixed at 0.5 for all experiments, following \cite{zhang2024heteroscedastic,seitzer2022on}. Experiments were conducted using Adam Optimizer \cite{kingma2014adam} on an NVIDIA A100 GPU with a batch size of 1 for 3D experiments and 8 for 2D experiments. The learning rates are set as default values from official implementations for all baselines, and are set to be $1\times10^{-4}$ for AC-CAR and CAR.

\begin{table*}[!t]
\setlength{\tabcolsep}{1.5 mm}
\caption{Quantitative Results on the IXI datasets. * represents for values that AC-CAR significantly outperformed with $p$-value $< 0.01$ in a paired $t$-test. The running time for model training (hr) and the inference time per volume pair (sec) are reported.}
\centering
\scalebox{1.0}{
{
\begin{tabular}{ccccccccc}
  \toprule[1pt]
  \multirow{2}{*}{\textbf{Methods}} & \multicolumn{2}{c}{\textbf{T1w$\rightarrow$T2w}} & \multicolumn{2}{c}{\textbf{T1w$\rightarrow$PD}} & 
  \multicolumn{2}{c}
  {\textbf{T2w$\rightarrow$PD}}\\
    \cmidrule(lr){2-7}
  
   	& Dice $\uparrow$ & $J_{<0}$\% $\downarrow$ & Dice $\uparrow$ & $J_{<0}$\% $\downarrow$ & Dice $\uparrow$ & $J_{<0}$\% $\downarrow$\\
	\cmidrule(lr){1-7}
 Unregistered & 0.685*$\pm$0.062 & - & 0.685*$\pm$0.062 & - & 0.685*$\pm$0.062 & -\\
 SyN & 0.782*$\pm$0.035 & 0.571*$\pm$0.118\% & 0.768*$\pm$0.038 & 0.481*$\pm$0.122\% & 0.744*$\pm$0.043 & 0.489*$\pm$0.125\%\\
    VXM-LNCC & 0.780*$\pm$0.036 & 0.460$\pm$0.082\% & 0.768*$\pm$0.039 & 0.560*$\pm$0.089\% & 0.756*$\pm$0.045 & 0.621*$\pm$0.116\%\\
    VXM-MIND & 0.774*$\pm$0.035 & 0.627*$\pm$0.110\% & 0.763*$\pm$0.036 & 0.513*$\pm$0.097\% & 0.751*$\pm$0.043 & 0.618*$\pm$0.112\%\\
    MIDIR & 0.777*$\pm$0.032 & 0.473$\pm$0.103\% & 0.766*$\pm$0.034 & 0.464$\pm$0.096\% & 0.753*$\pm$0.036 & \textbf{0.346}$\pm$0.075\%\\
    SM-Shape & 0.701*$\pm$0.044 & 0.459$\pm$0.118\% & 0.698*$\pm$0.045 & 0.483*$\pm$0.093\% & 0.700*$\pm$0.059 & 0.350$\pm$0.132\%\\
    CAR & 0.794*$\pm$0.035 & 0.489*$\pm$0.087\% & 0.787*$\pm$0.035 & 0.433$\pm$0.078\% & 0.782*$\pm$0.042 & 0.439$\pm$0.082\%\\
    OTMorph & 0.775*$\pm$0.034 & 0.833*$\pm$0.115\% & 0.761*$\pm$0.038 & 0.841*$\pm$0.110\% & 0.748*$\pm$0.041 & 0.939*$\pm$0.145\%\\
    UTSRMorph & 0.781*$\pm$0.034 &  0.569*$\pm$0.094\% & 0.763*$\pm$0.038 &  0.574*$\pm$0.091\% & 0.764*$\pm$0.042 &  0.620*$\pm$0.109\%\\
    \textbf{AC-CAR (Ours)} & \textbf{0.805}$\pm$0.036 & \textbf{0.438}$\pm$0.074\% & \textbf{0.796}$\pm$0.035 & \textbf{0.427}$\pm$0.079\% & \textbf{0.791}$\pm$0.041 & 0.409$\pm$0.097\%\\
	\bottomrule[1pt]
\end{tabular}}}
\label{table:quant_IXI}
\end{table*}

\subsection{Baseline Methods}

AC-CAR is first compared with a conventional iterative-based registration method, SyN \cite{avants2008symmetric}, with mutual information as the similarity metric and a default Gaussian smoothing of 3 and three scales with 180, 80, 40 iterations, respectively, following the settings of TransMorph \cite{chen2022transmorph}. AC-CAR is also compared with SOTA learning-based methods, including VoxelMorph \cite{balakrishnan2019voxelmorph} using LNCC (VXM-LNCC) and MIND \cite{heinrich2012mind} (VXM-MIND), respectively, a multi-contrast mutual information-based registration approach MIDIR \cite{qiu2021learning}, a contrast-agnostic registration framework, SynthMorph \cite{hoffmann2021synthmorph}, and our previous work on contrast-agnostic registration, CAR \cite{wang2024car}, as well as two recently proposed methods, OTMorph \cite{10621700} and UTSRMorph \cite{10693635}. Synthetic images for training SynthMorph are generated from random label maps (SM-Shape). Note that the channel dimensions of all convolutional layers in VXM-LNCC, VXM-MIND, SM-Shape, and OTMorph are set to 64 for 3D experiments and 256 for 2D experiments to ensure a fair comparison with AC-CAR, and the channel dimension of CAR is the same as that of AC-CAR. In addition, we implemented the large version of UTSRMorph (UTSRMorph-large) for evaluation.

\section{Results}

\subsection{Comparison Studies on Registration Accuracy}\label{sec:regis_accuracy}

\subsubsection{3D Inter-patient Brain MRI Registration on CamCAN dataset}
The left side of Table.~\ref{table:quant_cam} summarizes the quantitative results for the 3D inter-patient registration task on the CamCAN dataset. It can be observed that AC-CAR achieved the best registration accuracy while maintaining a low folding ratio compared to other learning-based approaches. Compared to MIDIR, OTMorph, and CAR, AC-CAR achieves registration accuracy improvements of 4.7\%, 3.3\%, and 2.4\% in terms of Dice score, respectively. It should be noted that both AC-CAR and CAR only used T1w images for training, which demonstrates their superior generalization ability. Removing NCC during the training of AC-CAR results in performance degradation, demonstrating the effectiveness of adding NCC as the similarity loss. The improvement in terms of the Dice score of AC-CAR compared to CAR shows the effectiveness of our proposed ACFM module. Both AC-CAR and CAR significantly outperform the other contrast-agnostic registration approaches, SM-shape. Because it relies on randomly generated labels for training, and may not register well in terms of fine-grained details of the human brain. Fig.~\ref{figure 4.1} visualizes the registration error maps and deformation fields for a specific image pair, comparing AC-CAR with other baseline methods. We can observe from the figure that AC-CAR can achieve lower registration error and a smoother deformation field compared to other baselines.

\begin{table*}[!t]
\setlength{\tabcolsep}{1.5 mm}
\caption{Quantitative results of different contrast combinations on the 3D CamCAN dataset. * represents for values that AC-CAR significantly outperformed with $p$-value $< 0.01$ in a paired $t$-test.}
\centering
\scalebox{0.97}{
{
\begin{tabular}{ccccccccc}
  \toprule[1pt]
  \multirow{2}{*}{\textbf{Methods}} & \multicolumn{2}{c}{\textbf{T2w$\rightarrow$T1w}} & \multicolumn{2}{c}{\textbf{T1w$\rightarrow$T1w}} & 
  \multicolumn{2}{c}
  {\textbf{T2w$\rightarrow$T2w}}\\
    \cmidrule(lr){2-7}
  
   	& Dice $\uparrow$ & $J_{<0}$\% $\downarrow$ & Dice $\uparrow$ & $J_{<0}$\% $\downarrow$ & Dice $\uparrow$ & $J_{<0}$\% $\downarrow$ \\
	\cmidrule(lr){1-7}
 Unregistered & 0.609*$\pm$0.078 & - & 0.609*$\pm$0.078 & - & 0.609*$\pm$0.078\\
 SyN & 0.764*$\pm$0.023 & 0.493$\pm$0.164\% & 0.823*$\pm$0.014 & 0.736*$\pm$0.091\% & 0.782*$\pm$0.019 & 0.695*$\pm$0.134\%\\
    VXM-LNCC & 0.762*$\pm$0.036 & 0.468$\pm$0.168\% & 0.597*$\pm$0.029 & \textbf{0.392}$\pm$0.091\% & 0.592*$\pm$0.028 & \textbf{0.327}$\pm$0.089\% \\
    VXM-MIND & 0.738*$\pm$0.038 & \textbf{0.389}$\pm$0.255\% & 0.626*$\pm$0.045 & 0.445$\pm$0.281\% & 0.617*$\pm$0.033 & 0.489*$\pm$0.168\%\\
    MIDIR & 0.720*$\pm$0.030 & 0.404$\pm$0.138\% & 0.557*$\pm$0.026 & 0.862*$\pm$0.146\% & 0.556*$\pm$0.024 & 0.973*$\pm$0.174\%\\
    SM-Shape & 0.698*$\pm$0.059 & 0.484$\pm$0.150\% & 0.706*$\pm$0.063 & 0.413$\pm$0.161\% & 0.694*$\pm$0.066 & 0.524*$\pm$0.217\%\\
    CAR & 0.803*$\pm$0.029 & 0.598*$\pm$0.164\% & 0.825*$\pm$0.036 & 0.544*$\pm$0.148\% & 0.788*$\pm$0.042 & 0.408$\pm$0.147\%\\
    OTMorph & 0.787*$\pm$0.029 & 0.745*$\pm$0.169\% & 0.547*$\pm$0.028 & 1.451*$\pm$0.131\% & 0.543*$\pm$0.027 & 1.601*$\pm$0.125\%\\
    UTSRMorph & 0.793*$\pm$0.026 &  0.463$\pm$0.100\% & 0.571*$\pm$0.025 &  0.857*$\pm$0.129\% & 0.558*$\pm$0.022 &  0.902*$\pm$0.159\%\\
    \textbf{AC-CAR (Ours)} & \textbf{0.814}$\pm$0.027 & 0.521$\pm$0.137\% & \textbf{0.844}$\pm$0.033 & 0.488$\pm$0.128\% & \textbf{0.799}$\pm$0.038 & 0.401$\pm$0.101\%\\
	\bottomrule[1pt]
\end{tabular}}}
\label{table:quant_comb}
\end{table*}

\subsubsection{Generalization Experiments on 3D Brain MRI registration on CamCAN dataset}

To assess the generalizability of AC-CAR, we evaluated its performance on the CamCAN dataset across different contrast combinations. Specifically, we tested T2w-to-T1w, T1w-to-T1w, and T2w-to-T2w registrations. For generalizability evaluation, we directly employed AC-CAR and other baseline models pretrained on the CamCAN dataset originally used for T1w-T2w registration. The results are shown in Table \ref{table:quant_comb}. We can observe from the table that AC-CAR still performs the best among all the methods. For the T2w-T1w registration task, AC-CAR achieves registration accuracy improvements of 2.7\% and 2.1\% compared to OTMorph, UTSRMorph in terms of Dice score, respectively. For both the T1w-to-T1w and T2w-to-T2w registration tasks, all methods except AC-CAR, CAR, and SM-shape exhibited substantial performance degradation. The Dice score of most baseline methods is even lower than the unregistered image. This is likely because these methods were trained specifically for the T1w-to-T2w registration task. While they could maintain performance on the related T2w-to-T1w task, they failed on the other two tasks due to the distributional shift between the training and test data. For SM-Shape, the performance showed little variation across the three tasks since it was trained on randomly generated label maps. However, the registration accuracy of SM-Shape remained well below that of CAR and AC-CAR. Furthermore, AC-CAR consistently outperformed CAR across all three tasks with statistical significance, demonstrating that the proposed ACFM module effectively enhances model generalizability.

\subsubsection{3D Inter-patient Brain MRI Registration on IXI dataset}
Table.~\ref{table:quant_IXI} summarizes the quantitative results for the 3D inter-patient registration task on the IXI dataset. We reported the registration accuracy in terms of Dice score and deformation regularity in terms of folding ratio for T1w-T2w, T1w-PD, and T2w-
PD registration tasks. It can be observed that AC-CAR still consistently achieved the best registration accuracy while maintaining a low folding ratio compared to other learning-based approaches for all three registration tasks. For the T1w-T2w and T1w-PD registration tasks, AC-CAR achieves Dice improvements of 2.3\% and 2.8\%, respectively, over the second-best baseline methods (excluding CAR). For the T2w-PD registration task, despite not being exposed to either contrast during training, AC-CAR demonstrates superior registration accuracy compared to all other methods, achieving Dice improvements of 2.7\% over the second-best baseline method (excluding CAR), demonstrating the superior generalizability of AC-CAR. Compared to CAR, AC-CAR also shows statistically significant improvement over Dice score, showing the effectiveness of our proposed ACFM module.

\subsubsection{2D cardiac MRI registration for T1 Mapping Sequence}
The right part of Table.~\ref{table:quant_cam} shows the quantitative comparison results of the 2D cardiac registration task for the T1 mapping sequence in the CMRxRecon dataset. AC-CAR still achieves the best registration accuracy and good deformation regularity. Compared to VXM-LNCC and UTSRMorph, AC-CAR still achieves a 3.3\% and 1.2\% higher Dice score, respectively. AC-CAR also outperforms its counterpart of removing NCC. Noted that AC-CAR and CAR still used only single contrast images for training, which further demonstrated the generalizability of AC-CAR when applied to eight other unseen contrasts during inference. OTMorph achieves slightly higher Dice scores but also higher HD95 values than AC-CAR when trained on all available contrasts. Nonetheless, AC-CAR demonstrates superior performance in delineating structural boundaries, as reflected by its lower HD95 in this setting. Fig.~\ref{figure 4.2} shows the qualitative registration results of a specific image pair of AC-CAR versus the baseline methods. It can be shown that AC-CAR can register more fine-grained details with smooth deformation fields. In addition, SM-Shape and OTMorph exhibit errors outside the cardiac region, despite the simulated deformation being confined within it. This may be because the leveraged image-to-image translation in OTMorph could introduce extra translation error, and the SM-shape could overlook details within the background region due to the Dice loss, resulting in background errors. On the contrary, AC-CAR and CAR use pre-augmented mono-contrast images to compute LNCC, which results in a zero dissimilarity difference for regions without deformation. This demonstrated that AC-CAR has a better capability to capture the deformation while using mono-contrast similarity for training.

\begin{table}[htbp]
\setlength{\tabcolsep}{1.5 mm}
\caption{Quantitative results of the generalizability experiment. The models are trained on CMRxRecon Cine data while tested on CMRxRecon T1 Mapping data. * represents for values that AC-CAR significantly outperformed with $p$-value $< 0.01$ in a paired $t$-test.}
\centering
\scalebox{0.87}{
\begin{tabular}{ccccccc}
  \toprule[1pt]
   	Methods & Dice $\uparrow$ & $|\nabla_{J}|$ $\downarrow$ & $J_{<0}$\% $\downarrow$ & HD95 $\downarrow$\\
	\cmidrule(lr){1-5}
 Unregistered & 0.783*$\pm$0.079 & - & - & 2.749*$\pm$0.546\\
    VXM-LNCC & 0.790*$\pm$0.129 & 0.089*$\pm$0.025 & 0.176*$\pm$0.168\% & 2.773*$\pm$1.200\\
    VXM-MIND & 0.835*$\pm$0.088 & 0.072*$\pm$0.021 & 0.010*$\pm$0.016\% & 2.233*$\pm$0.648\\
    MIDIR & 0.764*$\pm$0.047 & \textbf{0.013}$\pm$0.047 & 0.153*$\pm$0.125\% & 2.926*$\pm$0.927\\
    SM-Shape & 0.836*$\pm$0.078 & 0.049*$\pm$0.005 & 0.006$\pm$0.016\% & 2.302*$\pm$0.576 \\
    OTMorph & 0.865*$\pm$0.063 & 0.030*$\pm$0.008 & \textbf{0.004}$\pm$0.006\% & 2.464*$\pm$0.732 \\
    UTSRMorph & 0.848*$\pm$0.078 & 0.037*$\pm$0.017 & 0.005$\pm$0.009\% & 2.714*$\pm$1.032\\
    CAR & 0.866*$\pm$0.073 & 0.018$\pm$0.005 & 0.006$\pm$0.014\% & 1.915*$\pm$0.518\\
    \textbf{AC-CAR (Ours)} & \textbf{0.876}$\pm$0.062 & 0.019$\pm$0.006 & 0.006$\pm$0.015\% & \textbf{1.762}$\pm$0.506\\
	\bottomrule[1pt]
\end{tabular}}
\label{table:quant_gene}
\end{table}

\subsubsection{Generalization Experiments on 2D cardiac MRI registration}
To further evaluate the generalizability of AC-CAR, we also trained our model and baseline methods on CMRxRecon Cine MRI data and evaluated model performance on the multi-contrast T1 Mapping data (also registering misaligned images of $\mathrm{TI_{i=2:9}}$ to $\mathrm{TI_{1}}$). The results are shown in Table.~\ref{table:quant_gene}. AC-CAR remains to achieve the best registration accuracy and a 4\% higher Dice score compared to SM-Shape. It should be noted that AC-CAR and CAR achieve similar performance compared to their counterpart trained using the T1 Mapping sequence. This demonstrated that our proposed contrast-agnostic registration framework can generalize to unseen contrasts during inference without sacrificing performance. However, VXM-LNCC, VXM-MIND, MIDIR, OTMorph, and UTSRMorph suffered different degrees of performance degradation compared to using the T1 Mapping sequence for training. Compared with OTMorph and UTSRMorph, AC-CAR demonstrates a substantially greater improvement in HD95 than in Dice score. This demonstrated that AC-CAR can better deal with structural boundaries. SM-shape maintains the same performance since it was trained on randomly generated labels. Compared to CAR, AC-CAR still achieves a 1\% higher Dice score, demonstrating that the proposed ACFM can also further improve the network's generalizability.

\subsection{Evaluation on Registration Uncertainty}
\begin{figure}[htbp]
	\centering
\includegraphics[width=0.49\textwidth]{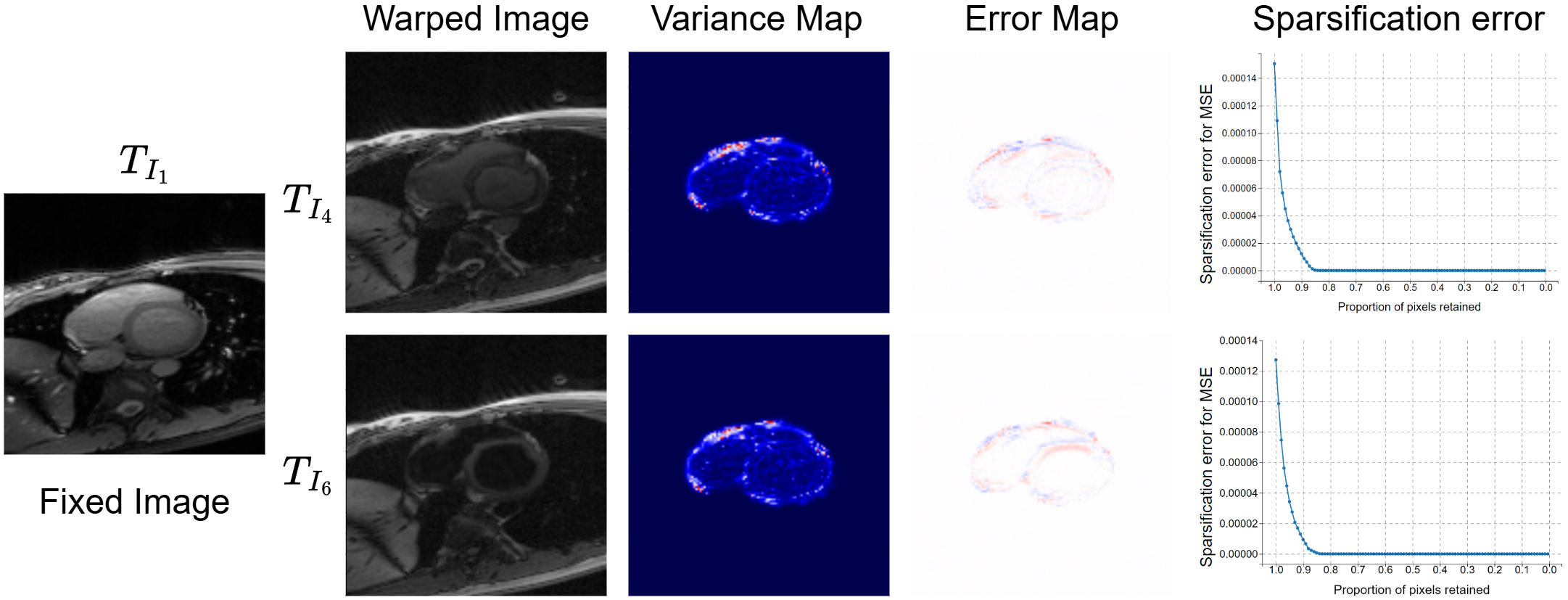} 
	\caption{Illustration of the estimated variance map and corresponding sparsification plot on the CMRxRecon dataset.} 
	\label{figure 4.3}
\end{figure}

We evaluate the estimated variance map through our proposed contrast-agnostic uncertainty estimation framework using the sparsification error plots from Poggi et al. \cite{poggi2020uncertainty} following Zhang et al. \cite{zhang2024heteroscedastic}. The sparsification error plot can provide a quantitative measure of the accuracy of the estimated variance. The plots are calculated by removing one pixel (voxel) at a time from the largest to the smallest variance magnitudes and measuring the MSE of the remaining pixels. An ideal sparsification error plot should decrease monotonically, indicating that the estimated uncertainty map can correctly identify pixels with the largest errors. As shown in Fig.~\ref{figure 4.3}, our proposed method can effectively estimate the registration uncertainty of the warped moving image as the sparsification plots are monotonically decreasing. The estimated variance is highly correlated with the registration error. This can be demonstrated by the high correlations between the areas with high variance in the variance map and the areas with high registration errors in the error map. 

\subsection{Evaluation on Feature Invariance}

\begin{figure}[htbp]
	\centering
\includegraphics[width=0.48\textwidth]{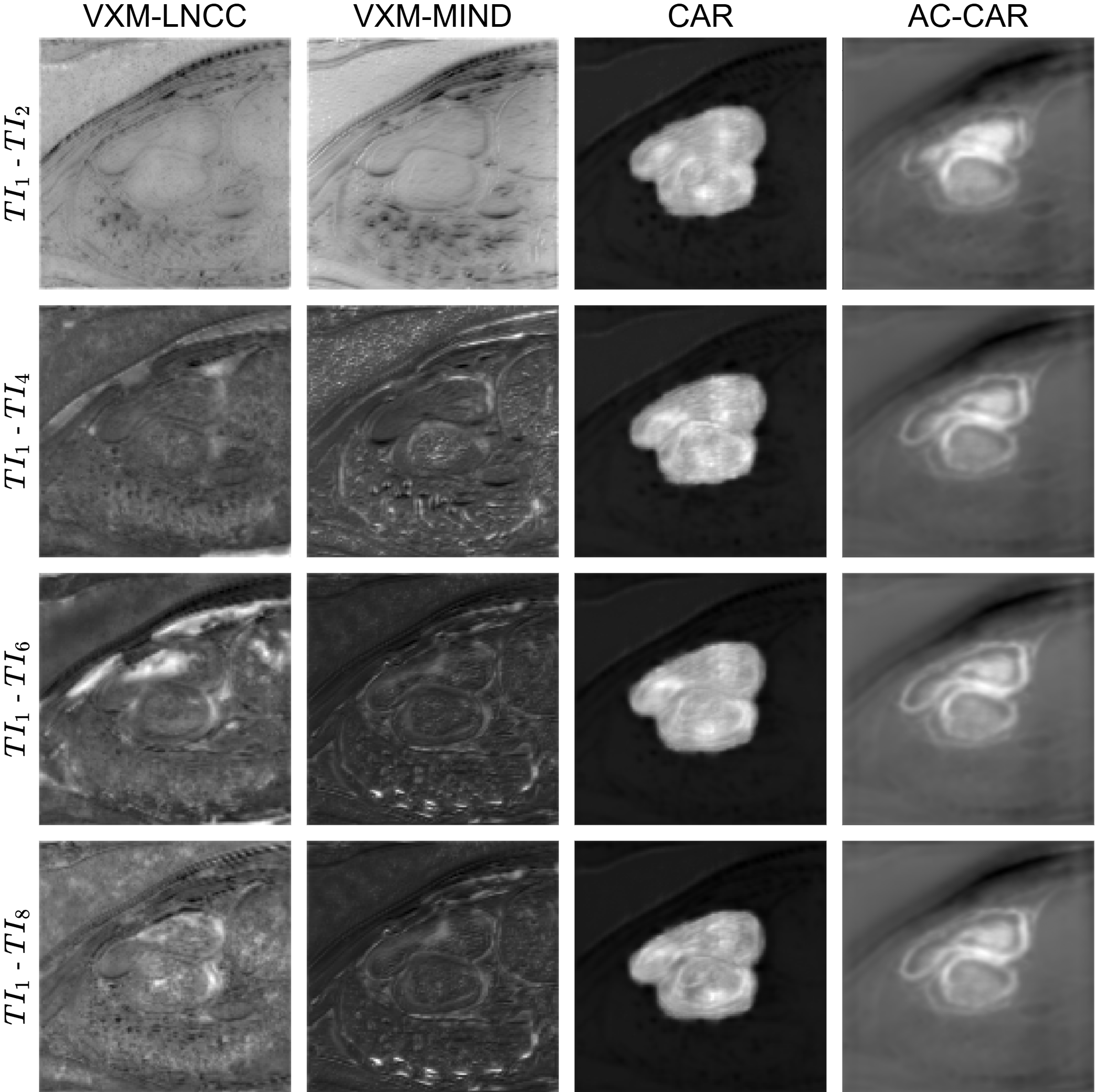} 
	\caption{Feature visualization from the final layer of the decoder of 4 feature pairs of the same subject on the CMRxRecon dataset.} 
	\label{figure 4.5}
\end{figure}

\begin{figure}[htbp]
	\centering
\includegraphics[width=0.35\textwidth]{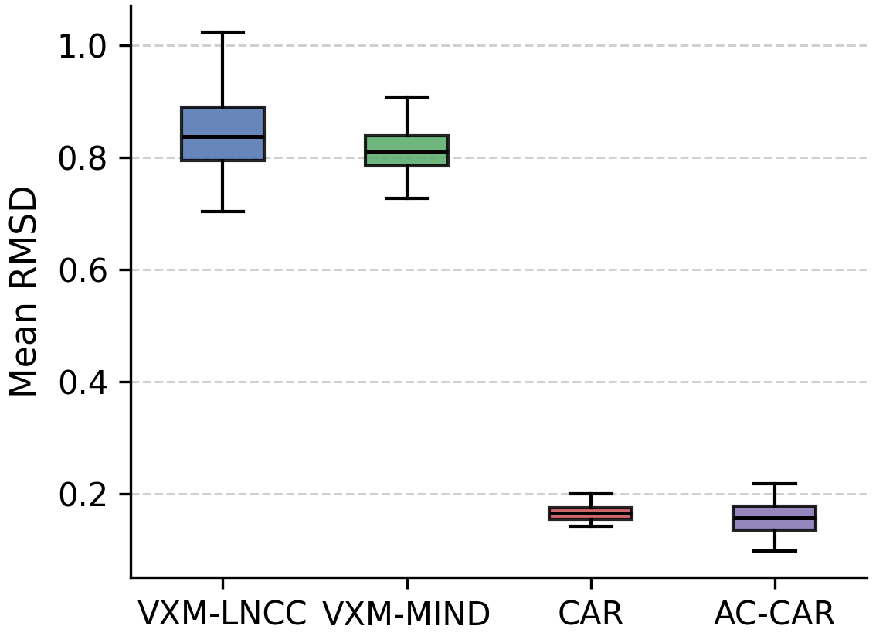} 
	\caption{Boxplot of Mean RMSD of 56 feature pairs across all subjects in the test set on the CMRxRecon dataset.} 
	\label{figure 4.6}
\end{figure}

To evaluate whether the features learned by AC-CAR are genuinely contrast-agnostic, we quantify the differences between features extracted from input image pairs that share the same anatomy but differ in imaging contrast. For each subject, we first extract features from eight input image pairs obtained from the T1 Mapping sequence without applying random spatial deformations ($\mathrm{TI_1-TI_{i=2:9}}$). Since the proposed CLR is applied at the end of the encoder, the learned features are highly abstract at that stage. Therefore, we utilize features from the final layer of the decoder, which maintains the same spatial resolution as the input images, to compute feature differences, following \cite{hoffmann2021synthmorph}. Specifically, we generate all 56 unique pairs via permutation of the eight extracted features and calculate the root-mean-square difference (RMSD) for each pair, following \cite{hoffmann2021synthmorph}. The mean RMSD across the 56 feature pairs is then computed to quantify the model’s feature invariance across the nine different imaging contrasts of a single subject. We compared AC-CAR with VXM-LNCC, VXM-MIND, and CAR, as all of them adopted a U-Net architecture. Figure \ref{figure 4.6} presents the quantitative results as a boxplot of the mean RMSD of the 56 feature pairs across all test subjects, while Figure \ref{figure 4.5} shows qualitative feature visualizations from the final decoder layer for four contrast pairs of the same subject. From Figure \ref{figure 4.5}, we observe that the features extracted by AC-CAR and CAR remain consistent across different imaging contrasts of the same subject, whereas those produced by VXM-LNCC and VXM-MIND exhibit substantial variations in intensity and texture. However, compared with AC-CAR, the features from CAR appear oversimplified and lack anatomical detail, whereas AC-CAR preserves fine-grained anatomical boundaries and detail. The quantitative results in Figure \ref{figure 4.6} further support these observations. AC-CAR achieves substantially lower mean RMSD compared to VXM-LNCC and VXM-MIND, demonstrating superior invariance to contrast variations. While CAR also achieves low RMSD values, it falls short of AC-CAR in registration accuracy according to previous experiments in Section \ref{sec:regis_accuracy}, highlighting the advantage of incorporating the proposed ACFM module, which enables the network to adaptively learn feature representations that are both anatomically informative and robust to variations in imaging contrast.

\subsection{Ablation Studies}

\subsubsection{Ablation study on network components}\label{sec:abl_nc}

We assess the effectiveness and computational cost of each module in AC-CAR by systematically removing them and evaluating the resulting model performance along with the associated training and inference times on the CamCAN dataset. The results are summarized in Table.~\ref{ablation: model}. CA denotes the contrast augmentation scheme in Section \ref{sec3.1}. UE denotes the uncertainty estimation module in Section \ref{sec3.4}. By removing the proposed ACFM, we remove the conditioning mechanism in Fig.\ref{figure 3.2} while the two convolution operations per layer are retained.

Table.~\ref{ablation: model} shows that removing either ACFM or CLR leads to a performance drop, which demonstrates that ACFM and CLR mutually reinforce contrast-invariant feature learning. Notably, removing ACFM results in a more significant drop than removing CLR. This suggests that CLR alone cannot adequately handle distribution shifts across arbitrary contrasts, as the network is unable to modulate itself and therefore tends to overfit to each contrast seen at training. Moreover, adding CLR on top of ACFM can further improve the model performance, which shows that CLR can be regarded as a guided constraint for the modulation in ACFM to further enhance contrast-invariant feature learning. Table.~\ref{ablation: model} also shows that incorporating UE into AC-CAR does not yield a statistically significant change in model performance. The UE module primarily serves to provide trustworthiness to the AC-CAR.

For training time, the addition of the proposed ACFM results in only a marginal increase, extending the total from 31 hours to 34 hours. The exclusion of CLR yields the shortest training time, as each iteration involves only a single forward pass. By contrast, the most substantial increase arises from the UE module, which requires training an auxiliary variance network to estimate the uncertainty map. At the inference stage, however, the runtimes of all model variants remain virtually identical, indicating that both the ACFM and the UE module impose negligible additional computational overhead.

\begin{table}[t!]
\setlength{\tabcolsep}{1.0 mm}
\caption{Ablation study on each module of AC-CAR on the 3D CamCAN dataset and corresponding training and inference time on GPU. * represents that AC-CAR (with 4 checkmarks) significantly outperformed with $p$-value $< 0.01$ in a paired $t$-test.}
\centering
\scalebox{0.92}{
{
\begin{tabular}{llllllll}
  \toprule[1pt]
	 CA & ACFM & CLR & UE & Dice $\uparrow$ & $J_{<0}\%$ $\downarrow$ & Training & Inference \\
   \cmidrule(lr){1-8}
    \checkmark & & \checkmark & &  0.793*$\pm$0.028 & 0.516$\pm$0.126\% & 31hr & 3.277s \\
     \checkmark & \checkmark & & &  0.798*$\pm$0.024 & 0.524$\pm$0.134\% & 30hr & 3.292s\\
      \checkmark & \checkmark & \checkmark &  & 0.807$\pm$0.027 & 0.517$\pm$0.126\% & 34hr & 3.324s\\
      \checkmark & \checkmark & \checkmark & \checkmark & \textbf{0.808}$\pm$0.026 & \textbf{0.494}$\pm$0.101\% & 46hr & 3.421s\\
    \bottomrule[1pt]
\end{tabular}}}
\label{ablation: model}
\end{table}

\subsubsection{Ablation study on the configuration of ACFM}
We first conduct the ablation study on using different frequency components as the condition for CIN. The ablation study in this section is conducted on the 2D slices of the CamCAN dataset for efficiency. The results are summarized in
Table.~\ref{ablation: LH}. \textbf{Harr-LL} means that we use the low-frequency decomposed image only as we proposed in Section \ref{sec3.2}. \textbf{Haar-H} means we used the other three decomposed images (LH, HL, HH) as the condition. \textbf{Haar-all} means we used all four decomposed images as a condition. It can be demonstrated that \textbf{Haar-all} suffers performance degradation by introducing other high-frequency components of the image for CIN. \textbf{Haar-H} experiences further performance degradation by discarding the low-frequency component of the image. This is because high-frequency components are more correlated with structural information than contrast information. This shows that the low-frequency component is more suitable for feature modulations to inversely remove contrast-related information through CIN. 

\begin{table}[!t]
\setlength{\tabcolsep}{1.0 mm}
\caption{Ablation Study on using different components of DWT decomposed images for ACFM on 2D slices on the CamCAN dataset. * represents for values that Haar-LL significantly outperformed with $p$-value $< 0.01$ in a paired $t$-test.}
\centering
\scalebox{0.92}{
\begin{tabular}{lllll}
  \toprule[1pt]
	 Variants & Dice $\uparrow$ & $|\nabla_{J}|$ $\downarrow$ & $J_{<0}\%$ $\downarrow$ & HD95 $\downarrow$ \\
   \cmidrule(lr){1-5}
   Haar-LL (Ours)& \textbf{0.751}$\pm$0.053 & \textbf{0.082}$\pm$0.010 & \textbf{0.498}$\pm$0.183\% & \textbf{2.949}$\pm$0.642 \\
    Haar-H & 0.733*$\pm$0.054 & 0.083$\pm$0.010 & 0.530*$\pm$0.186\% & 3.105*$\pm$0.659\\
     Haar-all &0.745*$\pm$0.057 & 0.084$\pm$0.010 & 0.551*$\pm$0.191\% & 3.015*$\pm$0.642\\
    \bottomrule[1pt]
\end{tabular}}
\label{ablation: LH}
\end{table}

In addition, we also performed an ablation study on different approaches for extracting low-frequency information from the original input image in ACFM, including various types of DWT or using FFT directly for low-pass filtering. \textbf{Daubechies} and \textbf{Biorthogonal} represent using Daubechies and Biorthogonal wavelets for extracting the low-frequency component of the image. \textbf{Haar-2nd}, \textbf{Haar-3rd}, and \textbf{Haar-4th} represent using the low-frequency image of the second, third, and fourth level of Haar DWT for all encoder layers, respectively. Table.~\ref{ablation: DWT} summarizes the results of different variants of our proposed method. It can be observed that all 3 types of DWT perform better than low-pass FFT. Among the three types of DWT, the Haar wavelet achieves the best performance while being the simplest and efficient window. Table.~\ref{ablation: DWT} also demonstrated that using a progressively increasing level of DWT to generate low-frequency images for each encoder layer outperforms using a fixed level of DWT
for all encoder layers.

\begin{table}[!t]
\setlength{\tabcolsep}{1.0 mm}
\caption{Ablation Study on different discrete wavelet transforms for ACFM on 2D slices on the CamCAN dataset. 
* represents for values that Haar significantly outperformed with $p$-value $< 0.01$ in a paired $t$-test.}
\centering
\scalebox{0.92}{
\begin{tabular}{lllll}
  \toprule[1pt]
	 Variants & Dice $\uparrow$ & $|\nabla_{J}|$ $\downarrow$ & $J_{<0}\%$ $\downarrow$ & HD95 $\downarrow$ \\
   \cmidrule(lr){1-5}
   Daubechies & 0.747$\pm$0.051 & 0.085*$\pm$0.008 & 0.548*$\pm$0.158\% & 2.992*$\pm$0.641 \\
    Biorthogonal & 0.746*$\pm$0.054 & 0.082$\pm$0.010 & \textbf{0.484}$\pm$0.186\% & 3.129*$\pm$0.659\\
     Low-pass FFT &0.746*$\pm$0.057 & 0.086*$\pm$0.010 & 0.556*$\pm$0.191\% & 2.998*$\pm$0.642\\
     Haar (Ours) & \textbf{0.751}$\pm$0.053 & 0.082$\pm$0.010 & 0.498$\pm$0.183\% & \textbf{2.949}$\pm$0.642\\
     Haar-2nd &0.748$\pm$0.053 & 0.083$\pm$0.010 & 0.528*$\pm$0.191\% & 2.969$\pm$0.665\\
     Haar-3rd &0.747$\pm$0.053 & 0.081$\pm$0.009 & 0.515*$\pm$0.182\% & 2.979$\pm$0.658\\
     Haar-4th &0.740*$\pm$0.061 & \textbf{0.080}$\pm$0.010 & \textbf{0.484}$\pm$0.164\% & 2.996*$\pm$0.654\\
    \bottomrule[1pt]
\end{tabular}}
\label{ablation: DWT}
\end{table}

\begin{figure}[htbp]
	\centering
\includegraphics[width=0.44\textwidth]{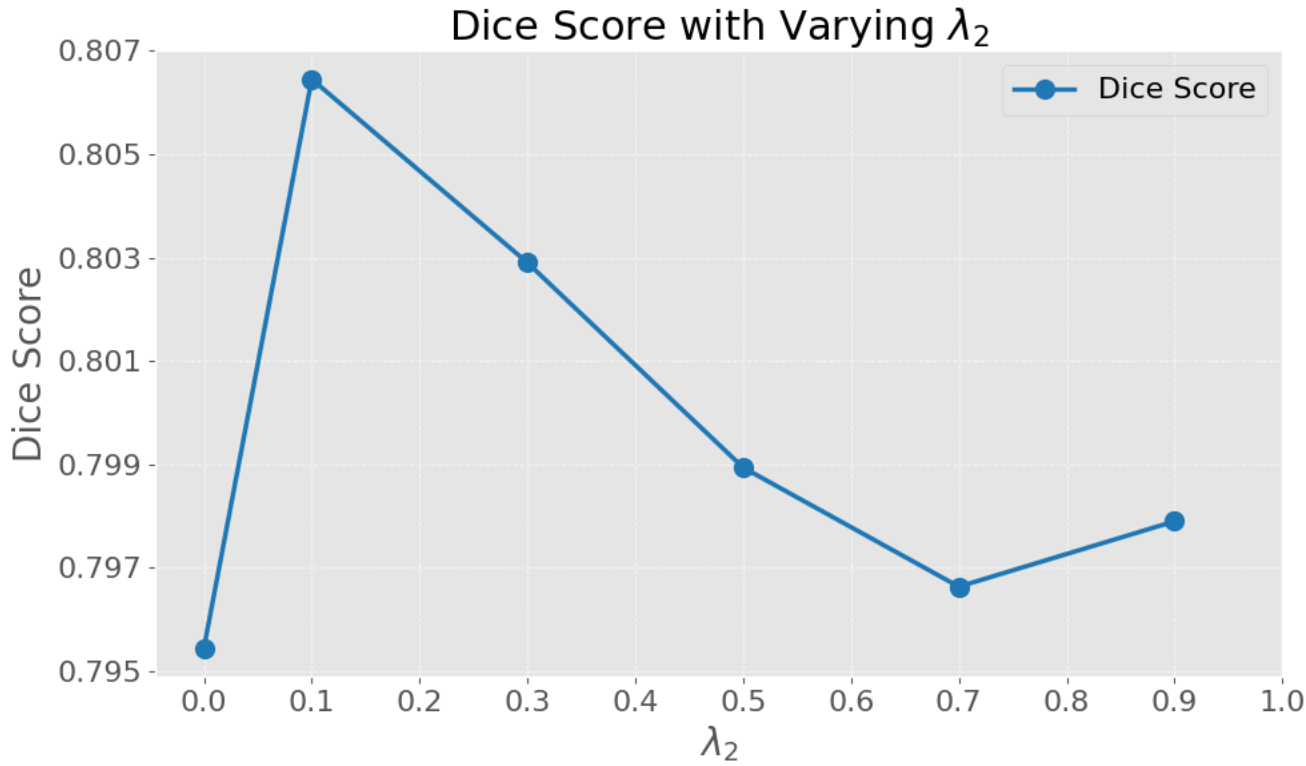} 
	\caption{Dice score on the 3D CamCAN dataset for AC-CAR  with varied contrast-invariance latent regularization parameters.} 
	\label{figure 4.4}
\end{figure}

\subsubsection{Ablation study on Contrast-Invariance Latent Regularization}
Fig.~\ref{figure 4.4} shows the registration accuracy in terms of the Dice score for different values of the contrast-invariance loss parameter $\lambda_2$ on the 3D brain MRI inter-patient registration task on the CamCAN dataset. It can be seen from the figure that when the CLR is removed at $\lambda_2$ equals $0$, the model performance suffers a severe degradation. The Dice score increases as the CLR parameter $\lambda_2$ increases to $0.1$, showing the effectiveness of our proposed Contrast-Invariant Latent Regularization (CLR). But when the CLR parameter $\lambda_2$ continues to increase, the Dice score experiences some degree of degradation and fluctuation. This demonstrated that excessive CLR can cause the network to overemphasize learning contrast-invariant features at the expense of information relevant to the registration task being ignored.

\begin{table}[!t]
\setlength{\tabcolsep}{1.5 mm}
\caption{Computational costs analysis for AC-CAR and baseline methods on the IXI dataset, reporting FLOPs, model parameters, and peak GPU memory measured during the inference stage, along with training and inference times.}
\centering
\scalebox{0.95}{
{
\begin{tabular}{cccccc}
  \toprule[1pt]
  \multirow{1}{*}{\textbf{Methods}} & \# Params & FLOPs & Memory & Training & Inference\\
    \cmidrule(lr){1-6}
    VXM-LNCC & 1.56M & 2204.49G & 9.18GB & 26hr & 2.478s\\
    VXM-MIND & 1.56M & 2204.49G & 9.18GB & 18hr & 2.568s\\
    MIDIR & 1.16M & 116.02G & 5.91GB & 20hr & 2.445s\\
    SM-Shape & 1.56M & 2204.49G & 9.00GB & 16hr & 2.613s\\
    CAR & 1.31M & 2161.97G & 8.87GB & 26hr & 2.517s\\
    OTMorph & 111.64M & 7405.47G & 10.84GB & 72hr & 2.308s\\
    UTSRMorph & 421.50M & 979.22G & 8.11GB & 34hr & 2.691s\\
    AC-CAR (w/o UE) & 2.83M & 2442.21G & 9.74GB & 31hr & 2.538s\\
    AC-CAR (Ours) & 4.05M & 4546.37G & 11.07GB & 43hr & 2.579s\\
	\bottomrule[1pt]
\end{tabular}}}
\label{table:comp_cost}
\end{table}

\subsection{Computational Cost Analysis}
To evaluate the computational cost of AC-CAR against the baseline learning-based approaches, we also reported the training and inference time, FLOPs, model parameters and peak GPU memory consumption during the inference stage in Table \ref{table:comp_cost}. We can observe that VXM-based approaches (VXM-LNCC, VXM-MIND, SM-Shape) share similar architectural complexity, each using around 1.56M parameters, over 2200 GFLOPs, and approximately 9GB of GPU memory during inference. MIDIR is the most lightweight baseline in terms of FLOPs and memory consumption, although its inference speed remains similar to other learning-based methods. In contrast, OTMorph and UTSRMorph impose substantially higher computational burdens, with OTMorph exhibiting the highest FLOPs and longest training time, and UTSRMorph employing the largest parameter count. Against these baselines, AC-CAR(w/o UE) represents AC-CAR without the uncertainty estimation module, which stands for the pure registration cost of our framework. With 2.83M parameters and 2442.21 GFLOPs, it remains comparable to VXM-based methods while maintaining a competitive inference time of 2.538s and modest memory usage (9.74GB). Compared with CAR, AC-CAR(w/o UE) exhibits an increased parameter count due to the ACFM module, yet it introduces only marginal increases in FLOPs, memory usage, and inference time. AC-CAR increases the computational cost due to the uncertainty estimation module, yet its inference time (2.579s) remains nearly identical to that of other deep learning approaches and comfortably within the narrow 2-3s range shared across all learning-based models. This indicates that the proposed ACFM and uncertainty estimation modules are lightweight, scalable addition that preserves the model’s deployment efficiency. AC-CAR therefore demonstrates superior accuracy and reliability while maintaining efficient inference, achieving a better trade-off between computational cost and performance.

\begin{table}[htbp]
\setlength{\tabcolsep}{1.0 mm}
\caption{Quantitative Results on the Learn2Reg datasets. * represents for values that AC-CAR significantly outperformed with $p$-value $< 0.01$ in a paired $t$-test.}
\centering
\scalebox{0.88}{
{
\begin{tabular}{ccccccc}
  \toprule[1pt]
   	Methods & Dice $\uparrow$ & $|\nabla_{J}|$ $\downarrow$ & $J_{<0}$\% $\downarrow$ & HD95 $\downarrow$\\
	\cmidrule(lr){1-5}
    Unregistered & 0.498*$\pm$0.127 & - & - & 13.817*$\pm$7.649\\
    VXM-MIND & \textbf{0.704}$\pm$0.120 & 0.058*$\pm$0.004 & \textbf{0.367}$\pm$0.109\% & \textbf{7.778}$\pm$3.088\\
    MIDIR & 0.635*$\pm$0.098 & 0.059*$\pm$0.005 & 0.589*$\pm$0.353\% & 11.656*$\pm$3.798\\
    SM-Shape & 0.605*$\pm$0.181 & 0.092*$\pm$0.010 & 0.398$\pm$0.065\% & 12.364*$\pm$8.647 \\
    \textbf{AC-CAR (Ours)} & 0.644$\pm$0.108 & \textbf{0.047}$\pm$0.005 & 0.463$\pm$0.160\% & 9.207$\pm$3.709\\
	\bottomrule[1pt]
\end{tabular}}}
\label{quant: L2R}
\end{table}

\subsection{Preliminary Evaluation on Multi-modal Registration}
\begin{figure}[htbp]
	\centering
\includegraphics[width=0.48\textwidth]{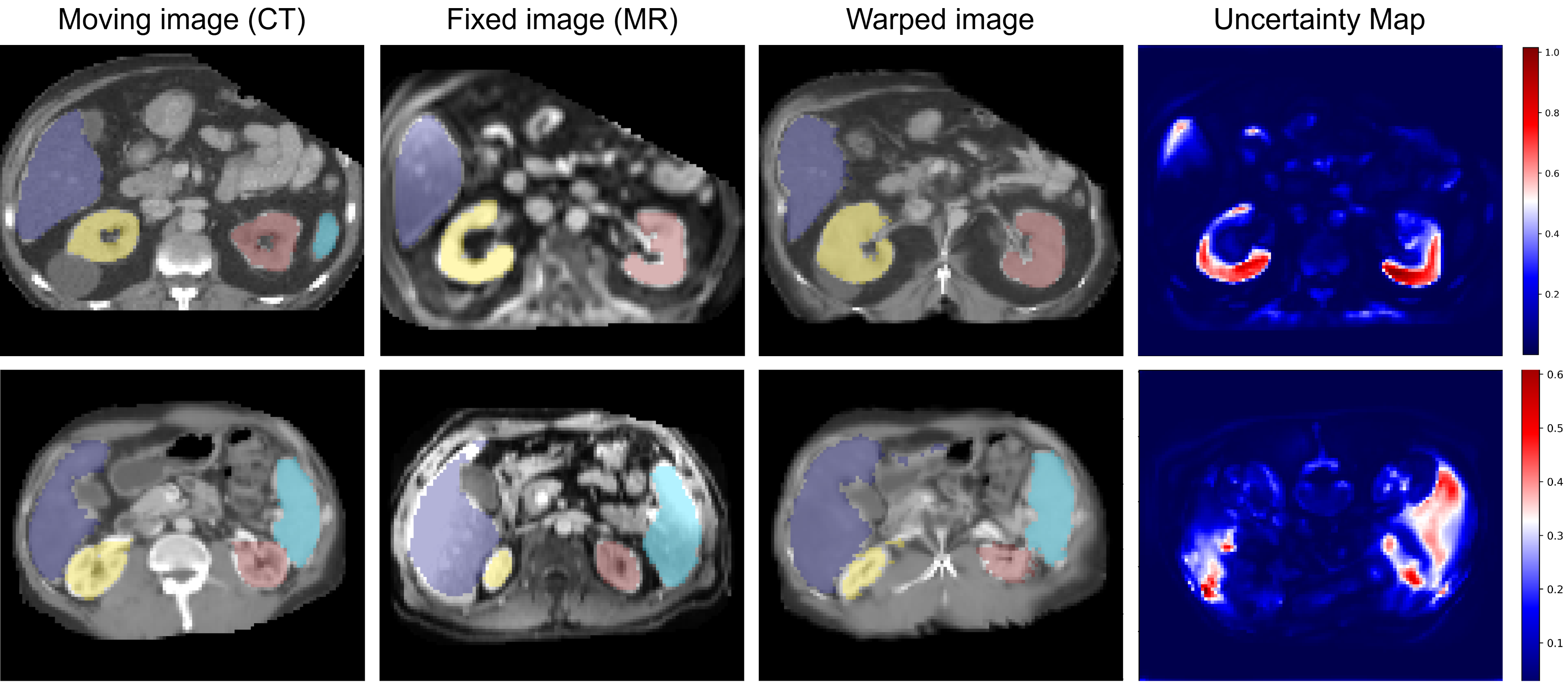} 
	\caption{Qualitative results of AC-CAR on CT–MR abdominal registration and the corresponding uncertainty maps for two test pairs from the Learn2Reg dataset.}
	\label{L2R_UE}
\end{figure}

To evaluate the generalizability of the proposed contrast-agnostic representation beyond multi-contrast scenarios, we further conduct preliminary experiments on a multi-modal abdominal CT-MR registration task using the Learn2Reg dataset \cite{hering2022learn2reg}. Table.~\ref{quant: L2R} summarizes the quantitative results. Although AC-CAR was not originally designed for multi-modal registration, the quantitative results indicate that it attains competitive performance across all evaluation metrics on the Learn2Reg dataset. Specifically, AC-CAR achieves the second-highest registration accuracy among the baseline methods, with a Dice score of 0.644 and an HD95 of 9.207, outperforming both MIDIR and SM-Shape. Notably, AC-CAR also yields the lowest gradient of Jacobian and a comparable folding ratio. These findings suggest that, despite the absence of an explicit multi-modal design, AC-CAR exhibits a substantial degree of intrinsic robustness to modality variability. This capability implies that the proposed contrast-agnostic representation has the potential to be extended toward multi-modal deformable registration. Figure \ref{L2R_UE} presents qualitative CT-MR registration results and the corresponding uncertainty maps for two test pairs. Note that no subject has paired and perfectly aligned CT-MR volumes, making dense error evaluation through error map or sparcification plot infeasible. We therefore only show the uncertainty maps. AC-CAR achieves visually accurate alignment, and the highlighted regions in the uncertainty maps generally correspond to areas of misregistration in the warped images. However, some misaligned regions are not fully captured by the uncertainty estimates, likely because the current contrast augmentation scheme can deal with modality variation to some extent but still has inherent limitations. This limitation will be further investigated in our future work.

\section{Discussion and Conclusion}\label{sec:conclusion}

We present a novel adaptive conditional contrast-agnostic deformable image registration framework (AC-CAR) that can register arbitrary contrasts of images without observing them during training. To achieve this, we propose a random convolution-based contrast augmentation scheme to simulate images with varying contrast as the input to our network. This prevents the network from overfitting a certain distribution of contrast to improve the network's generalizability. This has been demonstrated by the generalizability experiment in Table.~\ref{table:quant_gene}, where AC-CAR and CAR perform significantly better than all other baseline approaches. To learn contrast-invariant features, we propose an ACFM to enable the network to modulate the feature adaptively and a CLR to regularize the feature in the latent space. ACFM and CLR can mutually reinforce each other for contrast-invariant feature learning. Meanwhile, we also showed that AC-CAR can provide uncertainty estimation for multi-contrast image registration. Experiments demonstrated that on both 3D inter-subject brain MRI registration and 2D intra-subject cardiac MRI registration tasks, AC-CAR outperformed state-of-the-art methods and exhibited superior generalizability. The estimated uncertainty is highly correlated with the registration error, improving the reliability of AC-CAR.

However, one limitation of this work is that ACFM uses 2D slices for 3D conditioning to improve computational efficiency. Although this design achieves promising results, it may not fully capture the complete volumetric contrast variations present in 3D images. The other limitation of this work relies on the existing contrast augmentation scheme, which only simulates variations between different contrasts and may not capture variations across multiple modalities (e.g., MR-CT or US-CT registration tasks). In future work, we aim to adapt our framework for multi-modal registration.

\bibliographystyle{ieeetr}
\bibliography{ref}

\begin{thebibliography}{10}

\bibitem{alam2018medical}
F.~Alam, S.~U. Rahman, S.~Ullah, and K.~Gulati, ``Medical image registration in image guided surgery: Issues, challenges and research opportunities,'' {\em Biocybernetics and Biomedical Engineering}, vol.~38, no.~1, pp.~71--89, 2018.

\bibitem{sotiras2013deformable}
A.~Sotiras, C.~Davatzikos, and N.~Paragios, ``Deformable medical image registration: A survey,'' {\em IEEE transactions on medical imaging}, vol.~32, no.~7, pp.~1153--1190, 2013.

\bibitem{viola1997alignment}
P.~Viola and W.~M. Wells~III, ``Alignment by maximization of mutual information,'' {\em International journal of computer vision}, vol.~24, no.~2, pp.~137--154, 1997.

\bibitem{maes1997multimodality}
F.~Maes, A.~Collignon, D.~Vandermeulen, G.~Marchal, and P.~Suetens, ``Multimodality image registration by maximization of mutual information,'' {\em IEEE transactions on Medical Imaging}, vol.~16, no.~2, pp.~187--198, 1997.

\bibitem{heinrich2012mind}
M.~P. Heinrich, M.~Jenkinson, M.~Bhushan, T.~Matin, F.~V. Gleeson, M.~Brady, and J.~A. Schnabel, ``Mind: Modality independent neighbourhood descriptor for multi-modal deformable registration,'' {\em Medical image analysis}, vol.~16, no.~7, pp.~1423--1435, 2012.

\bibitem{heinrich2013towards}
M.~P. Heinrich, M.~Jenkinson, B.~W. Papie{\.z}, S.~M. Brady, and J.~A. Schnabel, ``Towards realtime multimodal fusion for image-guided interventions using self-similarities,'' in {\em Medical Image Computing and Computer-Assisted Intervention--MICCAI 2013: 16th International Conference, Nagoya, Japan, September 22-26, 2013, Proceedings, Part I 16}, pp.~187--194, Springer, 2013.

\bibitem{cheng2024winet}
X.~Cheng, X.~Jia, W.~Lu, Q.~Li, L.~Shen, A.~Krull, and J.~Duan, ``Winet: Wavelet-based incremental learning for efficient medical image registration,'' in {\em International Conference on Medical Image Computing and Computer-Assisted Intervention}, pp.~761--771, Springer, 2024.

\bibitem{qin2018joint}
C.~Qin, W.~Bai, J.~Schlemper, S.~E. Petersen, S.~K. Piechnik, S.~Neubauer, and D.~Rueckert, ``Joint learning of motion estimation and segmentation for cardiac mr image sequences,'' in {\em Medical Image Computing and Computer Assisted Intervention--MICCAI 2018: 21st International Conference, Granada, Spain, September 16-20, 2018, Proceedings, Part II 11}, pp.~472--480, Springer, 2018.

\bibitem{qin2019unsupervised}
C.~Qin, B.~Shi, R.~Liao, T.~Mansi, D.~Rueckert, and A.~Kamen, ``Unsupervised deformable registration for multi-modal images via disentangled representations,'' in {\em International Conference on Information Processing in Medical Imaging}, pp.~249--261, Springer, 2019.

\bibitem{dey2022contrareg}
N.~Dey, J.~Schlemper, S.~S.~M. Salehi, B.~Zhou, G.~Gerig, and M.~Sofka, ``Contrareg: Contrastive learning of multi-modality unsupervised deformable image registration,'' in {\em International Conference on Medical Image Computing and Computer-Assisted Intervention}, pp.~66--77, Springer, 2022.

\bibitem{qiu2021learning}
H.~Qiu, C.~Qin, A.~Schuh, K.~Hammernik, and D.~Rueckert, ``Learning diffeomorphic and modality-invariant registration using b-splines,'' in {\em Medical Imaging with Deep Learning}, 2021.

\bibitem{Arar_2020_CVPR}
M.~Arar, Y.~Ginger, D.~Danon, A.~H. Bermano, and D.~Cohen-Or, ``Unsupervised multi-modal image registration via geometry preserving image-to-image translation,'' in {\em Proceedings of the IEEE/CVF Conference on Computer Vision and Pattern Recognition (CVPR)}, June 2020.

\bibitem{hoffmann2021synthmorph}
M.~Hoffmann, B.~Billot, D.~N. Greve, J.~E. Iglesias, B.~Fischl, and A.~V. Dalca, ``Synthmorph: learning contrast-invariant registration without acquired images,'' {\em IEEE transactions on medical imaging}, vol.~41, no.~3, pp.~543--558, 2021.

\bibitem{sideri2023mad}
V.~Sideri-Lampretsa, V.~A. Zimmer, H.~Qiu, G.~Kaissis, and D.~Rueckert, ``Mad: Modality agnostic distance measure for image registration,'' in {\em International Conference on Medical Image Computing and Computer-Assisted Intervention}, pp.~147--156, Springer, 2023.

\bibitem{ronchetti2023disa}
M.~Ronchetti, W.~Wein, N.~Navab, O.~Zettinig, and R.~Prevost, ``Disa: Differentiable similarity approximation for universal multimodal registration,'' in {\em International Conference on Medical Image Computing and Computer-Assisted Intervention}, pp.~761--770, Springer, 2023.

\bibitem{rueckert2000non}
D.~Rueckert, M.~J. Clarkson, D.~L. Hill, and D.~J. Hawkes, ``Non-rigid registration using higher-order mutual information,'' in {\em Medical Imaging 2000: Image Processing}, vol.~3979, pp.~438--447, SPIE, 2000.

\bibitem{pluim2000image}
J.~P. Pluim, J.~A. Maintz, and M.~A. Viergever, ``Image registration by maximization of combined mutual information and gradient information,'' in {\em Medical Image Computing and Computer-Assisted Intervention--MICCAI 2000: Third International Conference, Pittsburgh, PA, USA, October 11-14, 2000. Proceedings 3}, pp.~452--461, Springer, 2000.

\bibitem{zhuang2011nonrigid}
X.~Zhuang, S.~Arridge, D.~J. Hawkes, and S.~Ourselin, ``A nonrigid registration framework using spatially encoded mutual information and free-form deformations,'' {\em IEEE transactions on medical imaging}, vol.~30, no.~10, pp.~1819--1828, 2011.

\bibitem{mok2024modality}
T.~C. Mok, Z.~Li, Y.~Bai, J.~Zhang, W.~Liu, Y.-J. Zhou, K.~Yan, D.~Jin, Y.~Shi, X.~Yin, {\em et~al.}, ``Modality-agnostic structural image representation learning for deformable multi-modality medical image registration,'' in {\em Proceedings of the IEEE/CVF Conference on Computer Vision and Pattern Recognition}, pp.~11215--11225, 2024.

\bibitem{deng2023interpretable}
X.~Deng, E.~Liu, S.~Li, Y.~Duan, and M.~Xu, ``Interpretable multi-modal image registration network based on disentangled convolutional sparse coding,'' {\em IEEE Transactions on Image Processing}, vol.~32, pp.~1078--1091, 2023.

\bibitem{dice1945measures}
L.~R. Dice, ``Measures of the amount of ecologic association between species,'' {\em Ecology}, vol.~26, no.~3, pp.~297--302, 1945.

\bibitem{wang2024car}
Y.~Wang, S.~Du, S.~Zheng, X.~Luo, and C.~Qin, ``Car: Contrast-agnostic deformable medical image registration with contrast-invariant latent regularization,'' in {\em International Workshop on Biomedical Image Registration}, pp.~308--318, Springer, 2024.

\bibitem{risholm2013bayesian}
P.~Risholm, F.~Janoos, I.~Norton, A.~J. Golby, and W.~M. Wells~III, ``Bayesian characterization of uncertainty in intra-subject non-rigid registration,'' {\em Medical image analysis}, vol.~17, no.~5, pp.~538--555, 2013.

\bibitem{sedghi2019probabilistic}
A.~Sedghi, T.~Kapur, J.~Luo, P.~Mousavi, and W.~M. Wells, ``Probabilistic image registration via deep multi-class classification: characterizing uncertainty,'' in {\em Uncertainty for Safe Utilization of Machine Learning in Medical Imaging and Clinical Image-Based Procedures: First International Workshop, UNSURE 2019, and 8th International Workshop, CLIP 2019, Held in Conjunction with MICCAI 2019, Shenzhen, China, October 17, 2019, Proceedings 8}, pp.~12--22, Springer, 2019.

\bibitem{dalca2019unsupervised}
A.~V. Dalca, G.~Balakrishnan, J.~Guttag, and M.~R. Sabuncu, ``Unsupervised learning of probabilistic diffeomorphic registration for images and surfaces,'' {\em Medical image analysis}, vol.~57, pp.~226--236, 2019.

\bibitem{chen2022transmorph}
J.~Chen, E.~C. Frey, Y.~He, W.~P. Segars, Y.~Li, and Y.~Du, ``Transmorph: Transformer for unsupervised medical image registration,'' {\em Medical image analysis}, vol.~82, p.~102615, 2022.

\bibitem{zhang2024heteroscedastic}
X.~Zhang, D.~H. Pak, S.~S. Ahn, X.~Li, C.~You, L.~H. Staib, A.~J. Sinusas, A.~Wong, and J.~S. Duncan, ``Heteroscedastic uncertainty estimation framework for unsupervised registration,'' in {\em International Conference on Medical Image Computing and Computer-Assisted Intervention}, pp.~651--661, Springer, 2024.

\bibitem{dumoulin2017a}
V.~Dumoulin, J.~Shlens, and M.~Kudlur, ``A learned representation for artistic style,'' in {\em International Conference on Learning Representations}, 2017.

\bibitem{achille2018emergence}
A.~Achille and S.~Soatto, ``Emergence of invariance and disentanglement in deep representations,'' {\em The Journal of Machine Learning Research}, vol.~19, no.~1, pp.~1947--1980, 2018.

\bibitem{seitzer2022on}
M.~Seitzer, A.~Tavakoli, D.~Antic, and G.~Martius, ``On the pitfalls of heteroscedastic uncertainty estimation with probabilistic neural networks,'' in {\em International Conference on Learning Representations}, 2022.

\bibitem{balakrishnan2019voxelmorph}
G.~Balakrishnan, A.~Zhao, M.~R. Sabuncu, J.~Guttag, and A.~V. Dalca, ``Voxelmorph: a learning framework for deformable medical image registration,'' {\em IEEE transactions on medical imaging}, vol.~38, no.~8, pp.~1788--1800, 2019.

\bibitem{he2016deep}
K.~He, X.~Zhang, S.~Ren, and J.~Sun, ``Deep residual learning for image recognition,'' in {\em Proceedings of the IEEE conference on computer vision and pattern recognition}, pp.~770--778, 2016.

\bibitem{shafto2014cambridge}
M.~A. Shafto, L.~K. Tyler, M.~Dixon, J.~R. Taylor, J.~B. Rowe, R.~Cusack, A.~J. Calder, W.~D. Marslen-Wilson, J.~Duncan, T.~Dalgleish, {\em et~al.}, ``The cambridge centre for ageing and neuroscience (cam-can) study protocol: a cross-sectional, lifespan, multidisciplinary examination of healthy cognitive ageing,'' {\em BMC neurology}, vol.~14, pp.~1--25, 2014.

\bibitem{taylor2017cambridge}
J.~R. Taylor, N.~Williams, R.~Cusack, T.~Auer, M.~A. Shafto, M.~Dixon, L.~K. Tyler, R.~N. Henson, {\em et~al.}, ``The cambridge centre for ageing and neuroscience (cam-can) data repository: Structural and functional mri, meg, and cognitive data from a cross-sectional adult lifespan sample,'' {\em neuroimage}, vol.~144, pp.~262--269, 2017.

\bibitem{fischl2012freesurfer}
B.~Fischl, ``Freesurfer,'' {\em Neuroimage}, vol.~62, no.~2, pp.~774--781, 2012.

\bibitem{lowekamp2013design}
B.~C. Lowekamp, D.~T. Chen, L.~Ib{\'a}{\~n}ez, and D.~Blezek, ``The design of simpleitk,'' {\em Frontiers in neuroinformatics}, vol.~7, p.~45, 2013.

\bibitem{ledig2015robust}
C.~Ledig, R.~A. Heckemann, A.~Hammers, J.~C. Lopez, V.~F. Newcombe, A.~Makropoulos, J.~L{\"o}tj{\"o}nen, D.~K. Menon, and D.~Rueckert, ``Robust whole-brain segmentation: application to traumatic brain injury,'' {\em Medical image analysis}, vol.~21, no.~1, pp.~40--58, 2015.

\bibitem{wang2021recommendation}
C.~Wang, Y.~Li, J.~Lv, J.~Jin, X.~Hu, X.~Kuang, W.~Chen, and H.~Wang, ``Recommendation for cardiac magnetic resonance imaging-based phenotypic study: imaging part,'' {\em Phenomics}, vol.~1, pp.~151--170, 2021.

\bibitem{wang2023cmrxrecon}
C.~Wang, J.~Lyu, S.~Wang, C.~Qin, K.~Guo, X.~Zhang, X.~Yu, Y.~Li, F.~Wang, J.~Jin, {\em et~al.}, ``Cmrxrecon: an open cardiac mri dataset for the competition of accelerated image reconstruction,'' {\em arXiv preprint arXiv:2309.10836}, 2023.

\bibitem{9965747}
X.~Luo and X.~Zhuang, ``$\mathcal {X}$-metric: An n-dimensional information-theoretic framework for groupwise registration and deep combined computing,'' {\em IEEE Transactions on Pattern Analysis and Machine Intelligence}, vol.~45, no.~7, pp.~9206--9224, 2023.

\bibitem{hering2022learn2reg}
A.~Hering, L.~Hansen, T.~C. Mok, A.~C. Chung, H.~Siebert, S.~H{\"a}ger, A.~Lange, S.~Kuckertz, S.~Heldmann, W.~Shao, {\em et~al.}, ``Learn2reg: comprehensive multi-task medical image registration challenge, dataset and evaluation in the era of deep learning,'' {\em IEEE Transactions on Medical Imaging}, vol.~42, no.~3, pp.~697--712, 2022.

\bibitem{kingma2014adam}
D.~P. Kingma and J.~Ba, ``Adam: A method for stochastic optimization,'' {\em arXiv preprint arXiv:1412.6980}, 2014.

\bibitem{avants2008symmetric}
B.~B. Avants, C.~L. Epstein, M.~Grossman, and J.~C. Gee, ``Symmetric diffeomorphic image registration with cross-correlation: evaluating automated labeling of elderly and neurodegenerative brain,'' {\em Medical image analysis}, vol.~12, no.~1, pp.~26--41, 2008.

\bibitem{10621700}
B.~Kim, Y.~Zhuang, T.~S. Mathai, and R.~M. Summers, ``Otmorph: Unsupervised multi-domain abdominal medical image registration using neural optimal transport,'' {\em IEEE Transactions on Medical Imaging}, vol.~44, no.~1, pp.~165--179, 2025.

\bibitem{10693635}
R.~Zhang, H.~Mo, J.~Wang, B.~Jie, Y.~He, N.~Jin, and L.~Zhu, ``Utsrmorph: A unified transformer and superresolution network for unsupervised medical image registration,'' {\em IEEE Transactions on Medical Imaging}, vol.~44, no.~2, pp.~891--902, 2025.

\bibitem{poggi2020uncertainty}
M.~Poggi, F.~Aleotti, F.~Tosi, and S.~Mattoccia, ``On the uncertainty of self-supervised monocular depth estimation,'' in {\em Proceedings of the IEEE/CVF conference on computer vision and pattern recognition}, pp.~3227--3237, 2020.

\end{thebibliography}

\end{document}